\documentclass[10pt,twocolumn,journal]{IEEEtran}
\usepackage{float}
\usepackage{latexsym}
\usepackage[hidelinks]{hyperref}
\usepackage{amsfonts}
\usepackage{amsbsy}
\usepackage{amssymb}
\usepackage{times}
\usepackage{graphicx}
\usepackage{setspace}
\usepackage{enumerate}
\usepackage[usenames]{color}
\usepackage[dvips]{pstcol}
\usepackage{epstopdf}
\usepackage[caption=false]{subfig}
\usepackage{cite}
\usepackage{amssymb}
\usepackage{amsfonts}
\usepackage{graphicx}
\usepackage{epsfig}
\usepackage{psfrag}
\usepackage{xcolor}
\usepackage{amsfonts, bm}
\usepackage{epstopdf}
\usepackage{cite}
\usepackage{color}
\usepackage{xcolor}
\usepackage{subfig}
\usepackage{verbatim}
\usepackage{multirow}
\usepackage{booktabs}
\usepackage{amsthm}
\usepackage{makecell}
\usepackage{units}
\usepackage[linesnumbered, ruled]{algorithm2e}
\usepackage{algpseudocode}
\usepackage{amsmath}

\usepackage[linesnumbered, ruled]{algorithm2e}
\usepackage{algpseudocode}
\usepackage{amsmath}

\IEEEoverridecommandlockouts

\begin{document}
	
	\title{Dynamic Bandwidth Allocation for Hybrid Event-RGB Transmission}
	\author{\IEEEauthorblockN{Pujing Yang, Guangyi Zhang, Yunlong Cai, Lei Yu, and Guanding Yu}
		\thanks{
			Part of this work \cite{EVJSCC} will be presented at the IEEE 101st Vehicular Technology Conference, Oslo, Norway, June 2025.
			
			P. Yang, G. Zhang, Y. Cai, and G. Yu are with the College of Information Science and Electronic Engineering, Zhejiang University, Hangzhou 310027, China (e-mail: yangpujing@zju.edu.cn; zhangguangyi@zju.edu.cn; ylcai@zju.edu.cn; yuguanding@zju.edu.cn).
			
			
			L. Yu is with the School of Artificial Intelligence, Wuhan University, Wuhan, China (email:  ly.wd@whu.edu.cn).}}
	\maketitle
	\vspace{-3.3em}

	\begin{abstract}
		Event cameras asynchronously capture pixel-level intensity changes with extremely low latency. They are increasingly used in conjunction with RGB cameras for a wide range of vision-related applications.
		However, a major challenge in these hybrid systems lies in the transmission of the large volume of triggered events and RGB images. 
		To address this, we propose a transmission scheme that retains efficient reconstruction performance of both sources while accomplishing real-time deblurring in parallel. 
		Conventional RGB cameras and event cameras typically capture the same scene in different ways, often resulting in significant redundant information across their outputs. To address this, we develop a joint event and image (E-I) transmission framework to eliminate redundancy and thereby optimize channel bandwidth utilization.
		Our approach employs Bayesian modeling and the information bottleneck method to disentangle the shared and domain-specific information within the E-I inputs. This disentangled information bottleneck framework ensures both the compactness and informativeness of extracted shared and domain-specific information. Moreover, it adaptively allocates transmission bandwidth based on scene dynamics, i.e., more symbols are allocated to events for dynamic details or to images for static information.
		Simulation results demonstrate that the proposed scheme not only achieves superior reconstruction quality compared to conventional systems but also delivers enhanced deblurring performance.
	\end{abstract}
	
	\begin{IEEEkeywords}
		Semantic communications, joint source-channel coding, deblurring, event camera, information bottleneck.                              
	\end{IEEEkeywords}
	
	\IEEEpeerreviewmaketitle
	\section{Introduction}
\begin{figure*}[t]
		\begin{centering}
			\includegraphics[width=1 \textwidth]{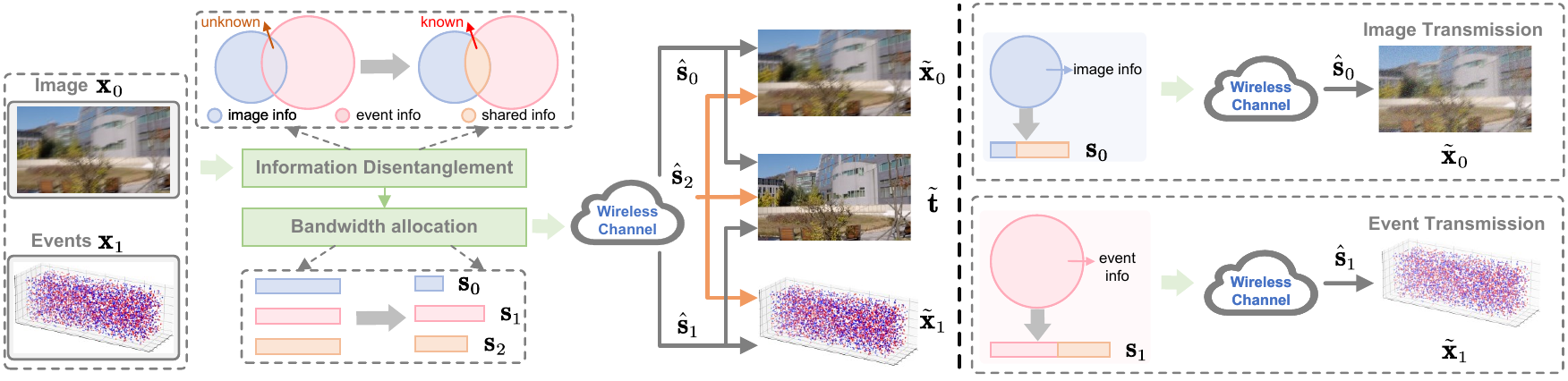}
			\par \end{centering}
		\caption{Comparison of our joint event-image transmission system (left) with the conventional system without considering the relevance between the E-I inputs (right). In our JEIT, the shared and domain-specific information is first disentangled. Then, channel bandwidth is allocated to each part based on its distribution. For example, the dynamics of the scene are relatively strong, and the captured image exhibits severe blurring; a larger proportion of symbols is allocated to events, which effectively capture dynamic scene changes. Notably, in the conventional system, the shared information is transmitted redundantly.}
		\label{framework}
	\end{figure*}

	Traditional RGB cameras, which capture a full image at a fixed rate, inherently struggle with dynamic scenarios with high speed (\textit{eg}., a blurry image due to motion blur is shown in Fig. \ref{framework}).
	In contrast, event cameras are bio-inspired sensors that asynchronously record pixel-level intensity changes, offering a wide dynamic range and exceptionally high temporal resolution \cite{gallego2020event, glover2016event, yu2022learning}.
	This complementary nature has inspired a growing body of research that integrates events with RGB images to enhance image quality \cite{kim2024frequency, xu2021motion, liang2024towards}.
	In the real-world deployment of these applications based on events and images, computational tasks are typically processed by high-performance servers, while task requests primarily originate from edge devices \cite{ldmic}. This necessitates the design of end-to-end transmission for both events and images.
	
	While extensive research has been conducted on image transmission, studies focusing on event transmission remain relatively scarce, which significantly hinders the practical application of event-based methods.
	To efficiently support downstream tasks based on events and images, it is crucial to develop an effective transmission strategy. However, several critical challenges remain.
	\begin{itemize}
		\item Event cameras tend to generate a substantial volume of events within a short period, primarily due to their ultra-high temporal resolution (often on the microsecond scale) in response to intensity changes. According to the Address Event Representation (AER) protocol, each event is represented by $8$ bytes \cite{AERprotocol}, including a timestamp, pixel coordinates, and polarity. As a result, directly transmitting these events can lead to significant bandwidth consumption.
		\item Since many event-based approaches require both images and events as inputs, the transmission system must handle the joint transmission of these two data modalities. Most existing works focus on single-modality inputs and fail to account for the relevance between different data modalities \cite{maojin, yufei2025}, thereby limiting transmission efficiency.
	\end{itemize}
	
	Consequently, conventional separation scheme which employs source coding and channel coding for single-modality input often
	fails in this case, and it is in great demand to develop new
	methods to handle these issues.
	
	In this paper, we demonstrate that joint source-channel coding (JSCC) methods enable efficient joint transmission of events and images (E-I) and deblurring. Leveraging deep neural networks, JSCC can selectively extract task-relevant semantic information while effectively mitigating channel impairments, thereby improving communication efficiency under bandwidth constraints\cite{20254,  Mingyu_TCCN2022, guoxiansheng}. Therefore, we propose a \textbf{J}oint \textbf{E}vent-\textbf{I}mage \textbf{T}ransmission system (JEIT), targeting reconstruction quality within constrained channel bandwidth. 
	Recognizing the inherent correlation between E-I data, we extract and transmit both shared and domain-specific information, thereby eliminating redundant transmission. 
	Notably, beyond reconstruction task, our framework concurrently processes received data to generate deblurred images in parallel, significantly accelerating the deblurring pipeline to meet real-time application requirements.
	
	While JSCC methods have demonstrated remarkable performance across various source modalities \cite{Xiaojiao_IOTJ2024, 20253}, including text, speech, and images, most existing efforts focus primarily on single-modality input data. When dealing with multi-modality data, the authors in \cite{Guangyi_TCOM2024} processed different modalities (image/text/speech) through separate encoders.
	However, this separate architecture fails to fully exploit inter-modal correlations, potentially resulting in redundant transmission.
	

	To address this issue, we seek to obtain compact yet informative representations of both shared and domain-specific information. 
	To begin with, we build Bayesian networks to model the relationships among the E-I inputs, the extracted features, and the desired outputs.
	Next, we disentangle the shared and domain-specific information based on the information bottleneck principle and introduce a novel objective function. 
	To make this objective tractable, we apply variational inference, enabling stable optimization while ensuring that the extracted representations remain both compact and effective. 
	Notably, our proposed objective naturally enables dynamic bandwidth allocation across different feature domains: more transmission symbols are assigned to events to capture dynamic details, or to images to preserve static information,
	depending on scene dynamics.
	
	We summarize our main contributions as follows:
	\begin{itemize}
		\item[(1)] We propose JEIT, a joint transmission and deblurring scheme for multi-modality inputs involving E-I data. To the best of our knowledge, this is the first work to achieve joint E-I data transmission, marking a significant advancement in efficient multi-modality data transmission.
		
		\item[(2)] 
		We build Bayesian networks and employ the information bottleneck principle to disentangle shared and domain-specific information, ensuring that the extracted information is both compact and informative.
		
		
		\item[(3)] 
		We propose a novel objective, which optimizes both the reconstruction performance, deblurring performance and channel bandwidth. Moreover, it adaptively allocates transmission bandwidth based on scene dynamics.

	\end{itemize}
	
	\begin{table}[t]
		\centering
		\caption{Major notations.}
		\begin{tabular*}{1.0\columnwidth}{c m{0.75\columnwidth}}
			\toprule
			\textbf{Notation} & \textbf{Defination} \\  
			\midrule
			\midrule
			$\mathbf{x}_0, \mathbf{x}_1, \mathbf{t}$ & The input image, the unified representation of input event stream, and the clear image. \\
			$\mathbf{\tilde{x}}_0, \mathbf{\tilde{x}}_1, \mathbf{\tilde{t}}$ & The reconstruction of input image, representation of event stream, and deblurred image.\\
			$\mathbf{s}_0, \mathbf{s}_1, \mathbf{s}_2$ & The transmitted image-specific, event-specific, and shared symbols.\\
			$\mathbf{\hat{s}}_0, \mathbf{\hat{s}}_1, \mathbf{\hat{s}}_2$ & The received image-specific, event-specific, and shared symbols. \\
			$\mathbf{x}_{0_{ab}}, \mathbf{t}_{ab}$ & The pixel of a blurry image and latent clear image at the location of $(a,b).$\\
			$T, \mathcal{T}, t, t_f$ & Exposure time, exposure interval, time between the exposure interval, and the midpoint of the exposure interval.\\
			$\mathbf{y}_0, \mathbf{y}_1, \mathbf{y}_2$ & The latent image-specific, event-specific, and shared information.\\
			$\mathbf{\hat{y}}_0, \mathbf{\hat{y}}_1, \mathbf{\hat{y}}_2$ & The quantized latent image-specific, event-specific, and shared information. \\
			$\mathbf{z}_0, \mathbf{z}_1, \mathbf{z}_2$ & The hyperprior variables for image-specific, event-specific, and shared information.\\
			$\mathbf{\hat{z}}_0, \mathbf{\hat{z}}_1, \mathbf{\hat{z}}_2$ & The quantized hyperprior variables for image-specific, event-specific, and shared information.\\
			$I^{\mathcal{G}}$ & The multi-information of a Bayesian network $\mathcal{G}$. \\
			$\mathcal{G}_{in}$ & The information compression network. \\
			$\mathcal{G}_{out}$ & The information preservation network. \\
			\bottomrule
		\end{tabular*}
		\label{tab:example_table}
	\end{table}

	\section{Prior Work}
	\subsection{Event-based Image Processing}
	Traditional RGB cameras are increasingly inadequate for meeting the requirements of high frame rates, low latency, and high dynamic range. In response to these challenges, event	cameras have emerged as a promising alternative, offering superior temporal resolution and low-latency capabilities for visual data acquisition.
	Early approaches for intensity reconstruction from events primarily leveraged the brightness constancy constraint \cite{cook2011interacting}, where the reconstruction process involved simultaneous optical flow estimation.
	In \cite{bardow2016simultaneous}, intensity reconstruction and optical flow estimation are formulated as a sliding window variational optimization framework.
	While event cameras provide information about intensity changes, RGB cameras deliver absolute intensity information.
	These complementary characteristics open up new opportunities for image reconstruction tasks, including image deblurring, denoising, and super-resolution \cite{continuous, pan2019bringing, sun2022event,wang2020event,ni2015visual}.
	A pioneering work in \cite{continuous} employed a complementary filter  to synthesize nearly continuous-time images by fusing events with blurry images. Later on, the authors in \cite{pan2019bringing} proposed a deblurring approach through the double integral of the events. Afterward, the authors in \cite{wang2020event} jointly considered deblurring, denoising, and super-resolution problems by exploiting a sparse learning network. Building upon this, in \cite{yu2023learning}, event noises were considered and  a rigorous event shuffle-and-merge scheme was proposed, leading to superior performance. Despite the superior performance of these event-based methods, the storage and transmission of this large volume of events remain challenging, limiting their practical application.
	
	
	\subsection{Semantic Communications}
	In recent years, the integration of deep learning models into wireless communication systems has given rise to a new paradigm: semantic communications.
	In contrast to the classical separation-based scheme using
	compression algorithms combined with the practical channel
	coding, semantic communications typically conceive of an
	integrated design of source coding and channel coding, also referred to as JSCC\cite{Huiqiang_TSP2021, Eirina_TCCN2019, zhang2025HJSCC, lamosc}.
	Semantic communications have shown significant effectiveness in processing various input modalities, such as text, speech, and images.
	A pioneering work on text transmission \cite{Huiqiang_TSP2021} demonstrated superior performance compared to conventional separation-based methods, highlighting the potential of semantic-aware design.
	Another notable advancement is DeepJSCC \cite{Eirina_TCCN2019}, which focuses on leveraging convolutional neural networks (CNNs) to transform source images into low-dimensional representations, offering strong resilience to channel noise.
	Building upon \cite{Eirina_TCCN2019}, researchers have incorporated more advanced techniques to further enhance performance.
	For instance, \cite{zhang2025HJSCC} introduced a progressive learned method to capture contextual information, achieving robust transmission against channel noise. 
	Additionally, \cite{lamosc} incorporated a large language model to extract textual features from images, significantly improving image reconstruction quality.
	Despite these advances, most existing efforts focus primarily on single-modality input data. When dealing with multi-modality data, such as events and RGB images, these approaches often fall short, as they do not fully exploit multi-modality relevance, limiting their efficiency and effectiveness.

	\section{Framework of Proposed JEIT}\label{SEC2}
	In dynamic scenes, event cameras are increasingly used in conjunction with RGB cameras for a wide range of vision-related applications. In this work, we specifically consider a new task, i.e., joint E-I data transmission and deblurring, in a hybrid camera system, consisting of a time-synchronized and spatially-aligned RGB camera and event camera. We first revisit the fundamental models for motion deblurring in such hybrid camera systems, then formulate the task of the joint E-I data transmission, for which we propose the JEIT model in the subsequent sections.
	
	\subsection{Motion Deblurring in Hybrid Camera Systems}
	Motion deblurring is a representative application using E-I data. In particular,
	a conventional RGB camera captures an image by integrating scene information over the exposure period: 
	$\mathbf{x}_{0_{ab}}=\frac{1}{T} \int_{t_f-T/2}^{t_f+T/2} \mathbf{t}_{ab}(t) d t$. In contrast, event cameras record intensity changes at each pixel asynchronously. Event stream triggered at pixel $(a, b)$ is defined as $e_{ab}(t) \triangleq  p \sum_{t_i \in \mathcal{T}}\delta(t-t_i)$, 
	whenever an event $ (a, b, t_i,p)$ occurs. The notations are defined in Table \ref{tab:example_table}. Based on this insight, the relationship between the captured image, event stream, and latent clear image can be modeled as: $\mathbf{x}_{0_{ab}} =\mathbf{t}_{ab}(t_f) \int_{t_f-T/2}^{t_f+T/2} \exp \left(c \int_{t_f}^{t} e_{ab}(s) d s\right) d t/T$. Therefore, it is possible to generate a deblurred image from the E-I input. 
	
	Since the amount of events triggered during a given period is inherently variable, it is hard to direct input event data into neural networks. Hence, a unified representation of events is of great necessity.
	Recent studies have shown the great benefits of using temporal data aggregation as the unified representation for event-based vision tasks \cite{time2020}. Inspired by \cite{sun2022event}, we divide the period $T$ into $2M$ intervals, defined by $2M+1$ temporal boundaries. We use $\mathbf{S}_m(a,b)$ to denote the integral of events from the  midpoint $t_f$ to the $m$-th boundary, for $m=0,1,...,2M$:
	\begin{equation}
		\mathbf{S}_m(a,b) = \int_{t_f}^{t_f+ (m-M)\frac{T}{2M}} e_{ab}(t) d t,
	\end{equation}	
	where for $m \textgreater M$, $\mathbf{S}_m(a,b)$ captures the accumulated positive events minus the negative events, while for $m \textless M$, it reflects accumulated negative events minus the positive events. This aggregation process transforms the asynchronous event stream into a fixed-size tensor $\mathbf{x}_1=[\mathbf{S}_0, \mathbf{S}_1,..., \mathbf{S}_{2M}]$ of dimensions $2M \times H \times W$, where $H \times W$ corresponds to the spatial resolution of the event camera. Here, $\mathbf{x}_1$ omits $\mathbf{S}_M$ as it's an all-zero tensor.


	\subsection{Joint E-I Data Transmission and Deblurring}
	In practical scenarios, event/RGB cameras are always deployed in the edge devices, while computational downstream tasks are usually processed by the high-performance servers. To this end, E-I data is usually delivered to the server via transmission link. However, existing methods only consider transmitting event data and RGB image data separately, which, however, overlook the data relevance between E-I data, and thus introduce abundant coding redundancy. Fortunately, this widely-used hybrid camera scenarios have naturally opened up directions for better system efficiency:
	\begin{itemize}
		\item In this hybrid systems, event and RGB cameras capture the same scenarios, and thus high information overlap exists. Hence, efficient joint coding mechanism is required to reduce the transmission redundancy.
		\item Both data streams are generated at the same device, making the joint signal processing available and efficient.
	\end{itemize}
	
	In general, this hybrid camera system generates two inputs for JEIT: the image $\mathbf{x}_0 \in \mathbb{R}^{n_0}$, where $n_0$ denotes the dimension of the input image, and the event stream $\mathbf{x}_1 \in \mathbb{R}^{n_1}$, where $n_1$ represents the size of the event representation. In particular, $n_0 = 3 \times H \times W$, while $n_1 = 2M \times H \times W$ after using temporal aggregation.
	
	\subsubsection{System Overview}
	Given the substantial overhead associated with independently transmitting E-I data, we conceive of a joint transmission system, and simultaneously generate deblurred images at the receiver. The JEIT framework consists of an image/event encoder, a shared encoder, an image/event decoder, and a deblurring decoder, as illustrated in Fig. \ref{framework}.
	The image encoder $f_{\bm{\theta}_0}: \mathbb{R}^{n_0}\rightarrow \mathbb{C}^{k_0}$ maps $\mathbf{x}_0$  to a complex vector $\mathbf{s}_0 \in \mathbb{C}^{k_0}$, where $\bm{\theta}_0$ denotes the parameters of the image encoder, and $k_0$ denotes the number of symbols to be transmitted. 
	The event encoder $f_{\bm{\theta}_1}: \mathbb{R}^{n_1}\rightarrow \mathbb{C}^{k_1}$ maps $\mathbf{x}_1$  to a complex vector $\mathbf{s}_1 \in \mathbb{C}^{k_1}$, where $\bm{\theta}_1$ denotes the parameters of the event encoder and $k_1$ denotes the output dimension.
	The shared encoder $f_{\bm{\theta}_2}: \mathbb{R}^{n_0+n_1}\rightarrow \mathbb{C}^{k_2}$, fuses features from both $\mathbf{x}_0$ and $\mathbf{x}_1$, and maps them to a complex vector $\mathbf{s}_2 \in \mathbb{C}^{k_2}$, where $\bm{\theta}_2$ denotes the paremeters of the shared encoder and $k_2$ denotes the number of transmitted symbols.
	Accordingly, the channel bandwidth ratio (CBR) is defined as $\rho = (k_0+k_1+k_2)/n_0$, denoting the average number of channel uses per image pixel. 
	In contrast to most existing transmission frameworks that employ a fixed CBR allocation for the input data, our approach dynamically allocates the CBR among image-specific, event-specific, and shared information components. This dynamic allocation strategy assigns more symbols to events in highly dynamic scenes while prioritizing images in static scenes.
	The encoding process is given by:
	\begin{equation}
		\mathbf{s}_0 = f_{\bm{\theta}_0}(\mathbf{x}_0),
		\mathbf{s}_1 = f_{\bm{\theta}_1}(\mathbf{x}_1),
		\mathbf{s}_2 = f_{\bm{\theta}_2}(\mathbf{x}_0, \mathbf{x}_1).
	\end{equation}
	
	Before transmission, we concatenate all encoders' outputs to obtain the transmitted symbol stream $\mathbf{s} = [\mathbf{s}_0, \mathbf{s}_1, \mathbf{s}_2]$. Then, we impose an average power constraint on $\mathbf{s}$. The channel input symbol stream $\mathbf{s}$ is then transmitted through a noisy wireless channel, modeled as $\mathcal{C}: \mathbb{C}^{k}\rightarrow\mathbb{C}^{k}$, where $k=k_0+k_1+k_2$. In this work, we consider an additive white Gaussian noise (AWGN) channel, and the transmission process is expressed as:
	\begin{equation}
		\mathbf{\hat{s}} \triangleq \mathcal{C}(\mathbf{s}) = \mathbf{s} + \mathbf{n},
	\end{equation}
	where $\mathbf{n} \sim \mathcal{CN}(\mathbf{0},\sigma^2 \mathbf{I})$ is a complex Gaussian vector with variance $\sigma^2$, and $\mathbf{I}$ denotes an identity matrix. 
	
	At the receiver, $\mathbf{\hat{s}}$ is first separated into $\mathbf{\hat{s}}_0, \mathbf{\hat{s}}_1$, and $\mathbf{\hat{s}_2}$. Then, the shared features $\mathbf{\hat{s}}_2$ combined with image-specific features $\mathbf{\hat{s}}_0$ are fed into the image decoder $g_{\bm{\phi}_0}:\mathbb{C}^{k_0+k_2} \rightarrow \mathbb{R}^{n_0}$ for original image reconstruction. Similarly, $\mathbf{\hat{s}}_2$ and $\mathbf{\hat{s}}_1$ are fed into the event decoder $g_{\bm{\phi}_1}:\mathbb{C}^{k_1+k_2} \rightarrow \mathbb{R}^{n_1}$ to reconstruct the event representation. For deblurred image $\mathbf{t} \in \mathbb{R}^{n_t}$, the shared features $\mathbf{\hat{s}}_2$ and the domain-specific features $\mathbf{\hat{s}}_0$ and $\mathbf{\hat{s}}_1$ are fed into deblurring decoder $g_{\bm{\phi}_t}:\mathbb{R}^{k_0+k_1+k_2} \rightarrow \mathbb{R}^{n_t}$ to perform deblurring. This decoding process is given by:
	\begin{equation}
		\mathbf{\tilde{x}}_0 = g_{\bm{\phi}_0}(\mathbf{\hat{s}}_0, \mathbf{\hat{s}}_2),
		\mathbf{\tilde{x}}_1 = g_{\bm{\phi}_1}(\mathbf{\hat{s}}_1, \mathbf{\hat{s}}_2),
		\mathbf{\tilde{t}} = g_{\bm{\phi}_t}(\mathbf{\hat{s}}_0, \mathbf{\hat{s}}_1, \mathbf{\hat{s}}_2).
	\end{equation}	
	
	Therefore, the optimization objective can be formulated as:
	\begin{equation} \label{objective}
		\begin{aligned}
			\bm{\Psi}^* &= \mathop{\arg\min}\limits_{\bm{\Psi}}\mathbb{E}_{p(\mathbf{x}_0,\mathbf{\tilde{x}}_0), p(\mathbf{x}_1,\mathbf{\tilde{x}}_1)} \\
			&[d(\mathbf{x}_0,\mathbf{\tilde{x}_0})+d(\mathbf{x}_0,\mathbf{\tilde{x}_0})+d(\mathbf{x}_0,\mathbf{\tilde{x}_0})\\
			&+\rho_0(\mathbf{x}_0, \mathbf{x}_1)+\rho_1(\mathbf{x}_0, \mathbf{x}_1)+\rho_2(\mathbf{x}_0, \mathbf{x}_1)],
		\end{aligned}
	\end{equation}
	where $\bm{\Psi}=\{\bm{\theta}_0, \bm{\theta}_1, \bm{\theta}_2, \bm{\phi}_0, \bm{\phi}_1, \bm{\phi}_2\}$, denoting all trainable parameters in the network. And we denote the CBR for image-specific, event-specific, and shared information as $\rho_0(\mathbf{x}_0, \mathbf{x}_1), \rho_1(\mathbf{x}_0, \mathbf{x}_1),$ and $\rho_2(\mathbf{x}_0, \mathbf{x}_1)$, where the trainable parameter set $\bm{\Psi}$ is omitted.
	
	\subsubsection{Challenges of JEIT}
	In general, our joint transmission framework exhibits better performance than separate transmission approaches. However, for the design of JEIT, we face the following problems:
	\begin{itemize}
		\item \textit{How to extract compact information?} Conventional separate transmission systems independently extract domain-specific information for source reconstruction, neglecting inter-modal correlations. This leads to redundant transmission of shared information, as shown in Fig. \ref{framework}. In contrast, our joint transmission scheme distinguishes the shared and domain-specific information through information disentanglement, ensuring the compactness and informativeness of the extracted features.
		\item \textit{How to realize dynamic CBR allocation?} In a separate transmission system, each pipeline independently optimizes its corresponding reconstruction, resulting in uncoordinated CBR allocation across modalities. In contrast, our proposed scheme leverages variational inference to reformulate the intractable disentanglement objective into a joint optimization of multi-modality distortion and CBR. This approach enables dynamic bandwidth allocation based on scene dynamics: more symbols are allocated to events in highly dynamic scenes (where they excel at capturing pixel-level changes), while static scenes prioritize image data.
	\end{itemize}
	In the following sections, we elaborate on how to realize these two goals in detail.

	\section{Methodology}\label{SEC3}
	To achieve compact information extraction, we propose an information disentanglement framework in Section \ref{ID}. Our approach first constructs Bayesian networks to model the relationships among the E-I inputs, extracted features, and target outputs. Guided by the information bottleneck principle, we then derive a novel objective function to effectively separate shared and domain-specific components while preserving essential information.
	The details of CBR allocation are presented in Section \ref{CBRallo} and \ref{CBRallo2}. Our key innovation lies in reformulating the intractable disentanglement objective into a tractable compression optimization problem. Building upon recent advances demonstrating compression algorithms’ effectiveness as CBR indicators for single-modality transmission \cite{Jincheng_JSAC2022}, we construct a multi-modality transmission framework from the compression model, enabling CBR allocation among the shared and domain-specific information.
	
	
	\subsection{Shared and Domain-Specific Information Disentanglement} \label{ID}
	\begin{figure}[t]
		\begin{centering}
			\includegraphics[width=0.46 \textwidth]{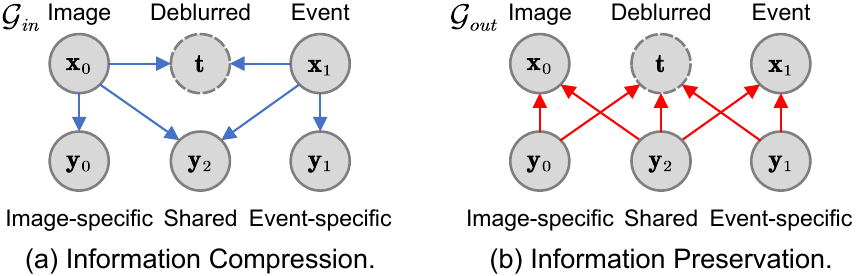}
			\caption{Bayesian networks for the base model, where the extracted information is not quantized. Here, blue lines represent the information compression process, while red lines indicate the information preservation process.}
			\label{Bayesian}
		\end{centering}
	\end{figure}
	
	In the hybrid camera systems considered in this work, RGB and event cameras capture the same scene from distinct perspectives. In this case, these two modalities inherently share some information, such as  object shapes and surface textures, while  also preserving domain-specific characteristics. Specifically, event cameras record intensity changes, which effectively capture motion information, whereas RGB cameras provide absolute pixel values that reflect  the scene’s color attributes. Directly transmitting them independently may lead to inefficient  channel bandwidth utilization due to redundant information.
	To enable more efficient transmission by minimizing redundancy, it is essential to distinguish and separately extract the shared and domain-specific information before transmission.	
	To deal with this issue, we propose a novel approach inspired by disentanglement learning \cite{Bayesian, multiinformation, multivariate2013}, which  aims to separate different feature representations within the data. This approach ensures that changes in one feature do not affect others, thus reducing inter-feature redundancy. Moreover, it enhances model interpretability by aligning each feature dimension with a distinct and meaningful semantic concept.
	
	\subsubsection{Modeling via Bayesian Networks}
	As shown in Fig. \ref{Bayesian}, our objective is to disentangle three parts from the input E-I data:  shared information, image-specific information, and event-specific information.
	Inspired by \cite{Bayesian}, we construct two pairwise Bayesian  networks: an information compression network $\mathcal{G}_{in}$ to promote the compactness of the representations, and an information preservation network $\mathcal{G}_{out}$ to ensure their informativeness. We refer to this model as the base model.
	
	Specifically, during the information compression stage, our goal is to disentangle compact  representations of image-specific information $\mathbf{y}_0$, event-specific information $\mathbf{y}_1$, and shared information $\mathbf{y}_2$ from the source variables $\mathbf{x}_0$ and $\mathbf{x}_1$. In this stage, $\mathbf{x}_0$ and $\mathbf{x}_1$ serve as the root nodes, while $\mathbf{y}_0, \mathbf{y}_1$, and $\mathbf{y}_2$ are  the leaf nodes, with the edges representing the compression relationships between them.
	In the information preservation stage, we aim to ensure that the domain-specific features $\mathbf{y}_0/\mathbf{y}_1$ preserve the discriminative characteristics of their respective inputs $\mathbf{x}_0/\mathbf{x}_1$, while the shared feature $\mathbf{y}_2$ retains the common information present in both modalities.
	The directed edges $\mathbf{y}_0/\mathbf{y}_1 \rightarrow \mathbf{x}_0/\mathbf{x}_1$ and $\mathbf{y}_2 \rightarrow \mathbf{x}_0/\mathbf{x}_1$ indicate the information preserved from the leaves back to the roots.
	Through the joint design of these pairwise Bayesian networks, the source variables are compressed into compact representations while retaining the relevant information necessary for reconstruction and deblurring tasks, thereby striking an effective balance between information compression and preservation.
	
	\begin{figure}[t]
		\begin{centering}
			\includegraphics[width=0.42 \textwidth]{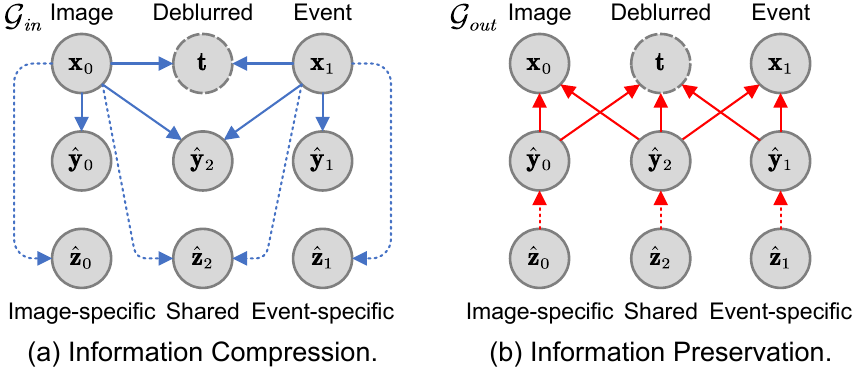}
			\caption{Bayesian networks for the base model and enhanced model. In the base model, the solid lines depict the relationships among all variables. In the enhanced model, the dotted lines represent additional relationships associated with hyperprior variables, while the solid lines retain the same structure and functionality as in the base model.}
			\label{Bayesian2}
		\end{centering}
	\end{figure}
	\subsubsection{Disentangling via Information Bottleneck}	
	As proposed in \cite{multiinformation}, the multi-information of a Bayesian network $\mathcal{G}$ quantifies the total amount of information shared among all variables $\mathbf{q}_1,...,\mathbf{q}_n$. It is defined by the \textit{Kullback–Liebler} divergence:
	\begin{equation}
		\!I^{\mathcal{G}}\!=\! D_{KL}[p(\mathbf{q}_1,...,\mathbf{q}_n)||p(\mathbf{q}_1)...p(\mathbf{q}_n)]\!\!=\!\!\! \sum_{\mathbf{q}_i\in \mathcal{G}} \!\!I(\mathbf{q}_i;\mathbf{P a}_{\mathbf{q}_i}^{\mathcal{G}}),
	\end{equation}
	where $\mathbf{P a}_{\mathbf{q}_i}^\mathcal{G}$ denotes the parents of node $\mathbf{q}_i$ in $\mathcal{G}$, and $I(\mathbf{q}_i;\mathbf{P a}_{\mathbf{q}_i}^\mathcal{G})$ is the mutual information between $\mathbf{P a}_{\mathbf{q}_i}^\mathcal{G}$ and $\mathbf{q}_i$. Mutual information measures the similarity between the joint distribution and the product of marginal distributions: greater similarity implies less information shared between nodes.
	\begin{itemize}
		\item Minimizing $I^{\mathcal{G}_{in}}$: As discussed in \cite{multivariate2013}, minimizing $I^{\mathcal{G}_{in}}$ ensures that $\mathbf{y}_0, \mathbf{y}_1$, and  $\mathbf{y}_2$ compactly and complementarily encode the information from $\mathbf{x}_0$ and $\mathbf{x}_1$, thereby maintaining their mutual disentanglement. This aligns with the desired information compression process.
		\item Maximizing $I^{\mathcal{G}_{out}}$: On the other hand, $I^{\mathcal{G}_{out}}$ reflects how much information the latent variables contain about the source variables. Since the goal is to predict each node from its parents in $\mathcal{G}_{out}$, we seek to maximize $I^{\mathcal{G}_{out}}$ to preserve informative representations. 
	\end{itemize}
	In line with the information bottleneck principle, the optimization objective is formulated as:
	\begin{equation} \label{original}
		\begin{aligned}
			&\mathop{\min}\limits_{\mathbf{q}_i \in \mathcal{G}}L_{}= -I^{\mathcal{G}_{out}} + \beta I^{\mathcal{G}_{in}} \\
			&=-\sum_{\mathbf{q}_i\in \mathcal{G}_{out}} I(\mathbf{q}_i;\mathbf{P a}_{\mathbf{q}_i}^{\mathcal{G}_{out}}) + \beta \sum_{\mathbf{q}_i\in \mathcal{G}_{in}} I(\mathbf{q}_i;\mathbf{P a}_{\mathbf{q}_i}^{\mathcal{G}_{in}}),
		\end{aligned}
	\end{equation}
	where $\beta \in (0,+\infty)$ is a Lagrange multiplier used to balance the trade-off between data compression and information preservation.
	Specifically, by incorporating the graphical structures of $\mathcal{G}_{in}$ and $\mathcal{G}_{out}$ from Fig. \ref{Bayesian} into Eq. \eqref{original}, the instantiated objective can be subsequently described as:
	\begin{align} \label{objective_base}
		&\mathop{\min}\limits_{\mathbf{y}_0, \mathbf{y}_1, \mathbf{y}_2}L_{}= \\
		&-\big[I(\mathbf{x}_0; \mathbf{y}_0, \mathbf{y}_2)
		+ I(\mathbf{x}_1; \mathbf{y}_1, \mathbf{y}_2) + I(\mathbf{t}; \mathbf{y}_0, \mathbf{y}_2, \mathbf{y}_1)\big] \notag \\
		&+ \beta \big[I(\mathbf{x}_0; \mathbf{y}_0)\!+\!I(\mathbf{x}_0, \mathbf{x}_1; \mathbf{y}_2) \!+ I(\mathbf{x}_1; \mathbf{y}_1) + I(\mathbf{t}; \mathbf{x}_0, \mathbf{x}_1)\big]. \notag
	\end{align}
	However, this objective poses challenges in estimating and optimizing the mutual information terms, as they typically involve integrals
	over high-dimensional spaces. To address this issue, we employ variational inference to derive tractable bounds for these terms, which will be discussed in the following subsections.
	\subsection{Optimization for Compression via Variational Inference}\label{CBRallo}
\begin{figure}[t]
		\begin{centering}
			\includegraphics[width=0.5 \textwidth]{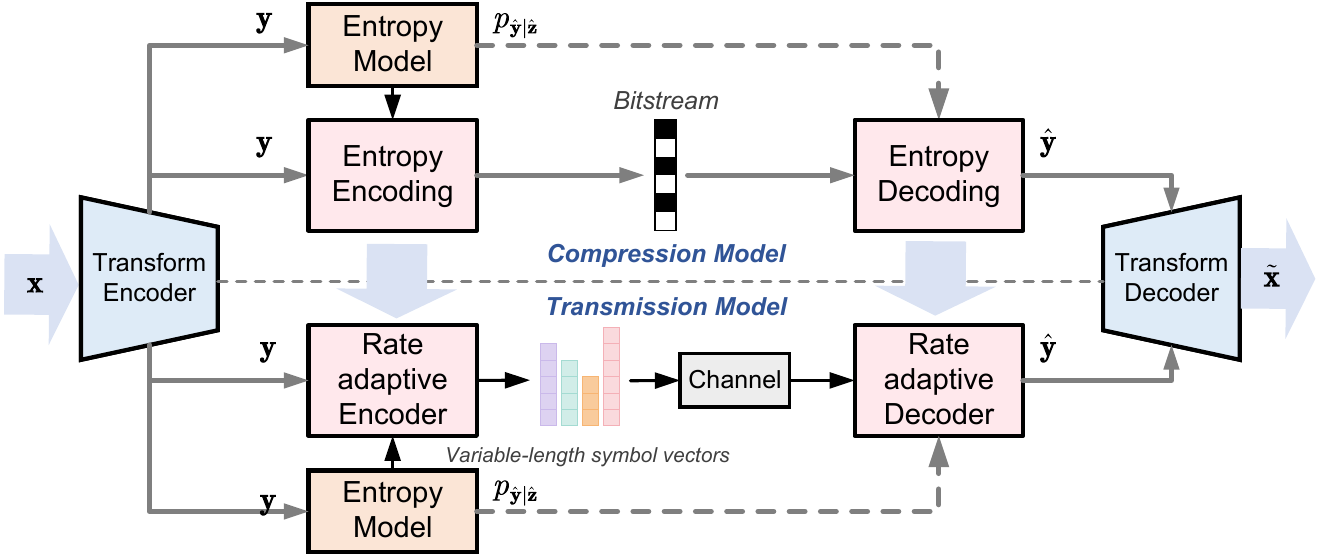}
			\caption{Diagram of a compression algorithm serving as a rate indicator for JSCC transmission process. The entropy model estimates the distribution of $\mathbf{y}$, based on which the bitstream length is determined.}
			\label{block}
		\end{centering}
	\end{figure}

	Intuitively, minimizing the mutual information between representations and inputs while maximizing that between representations and outputs in Eq. \eqref{objective_base} aligns with the fundamental rate-distortion trade-off in lossy compression, where the rate reflects the number of bits required to encode the input into a bitstream. 
	Compression algorithms can effectively serve as rate indicators within transmission frameworks \cite{Jincheng_JSAC2022}. 
	As illustrated in Fig. \ref{block}, in the compression process, a transform encoder first converts the source into a latent representation $\mathbf{y}$, and an entropy coding method is then utilized to encode $\mathbf{y}$ into a bitstream based on its distribution, which is estimated by an entropy model. Building upon this compression framework, \cite{Jincheng_JSAC2022} proposed a transmission model by replacing the entropy encoder with a rate-adaptive encoder to convert $\mathbf{y}$ into multiple vectors with different lengths. For clarity, this subsection first analyzes Eq. \eqref{objective_base} from the perspective of a compression problem in Section \ref{base}. Then we incorporate three hyperprior variables  to enhance the compression model in Section \ref{enhance}.
	\subsubsection{Base Bayesian Model Optimization} \label{base}
	To align with a compression scenario, we replace the representations $\mathbf{y}_0, \mathbf{y}_1,$ and $\mathbf{y}_2$ in the Bayesian network  with their quantized counterparts $\mathbf{\hat{y}}_0, \mathbf{\hat{y}}_1$, and $\mathbf{\hat{y}}_2$, as shown in Fig. \ref{Bayesian2}. The corresponding objective becomes:
	\begin{align} \label{objective_base2}
		&\mathop{\min}\limits_{\mathbf{\hat{y}}_0, \mathbf{\hat{y}}_1, \mathbf{\hat{y}}_2}L_{}= \\
		&-\big[I(\mathbf{x}_0; \mathbf{\hat{y}}_0, \mathbf{\hat{y}}_2)
		+ I(\mathbf{x}_1; \mathbf{\hat{y}}_1, \mathbf{\hat{y}}_2) + I(\mathbf{t}; \mathbf{\hat{y}}_0, \mathbf{\hat{y}}_2, \mathbf{\hat{y}}_1)\big] \notag \\
		&+ \beta \big[I(\mathbf{x}_0; \mathbf{\hat{y}}_0)\!+\!I(\mathbf{x}_0, \mathbf{x}_1; \mathbf{\hat{y}}_2) \!+ I(\mathbf{x}_1; \mathbf{\hat{y}}_1) + I(\mathbf{t}; \mathbf{x}_0, \mathbf{x}_1)\big]. \notag
	\end{align}
	
	To begin with, we note that the relationship between the deblurred image $\mathbf{t}$ and the sources $\mathbf{x}_0, \mathbf{x}_1$ is deterministic. Thus, $I(\mathbf{t}; \mathbf{x}_0, \mathbf{x}_1)$ can be omitted. To maximize $I(\mathbf{x}_0;\mathbf{\hat{y}}_0,\mathbf{\hat{y}}_2)$, we introduce a  variational distribution $q(\mathbf{x}_0|\mathbf{\hat{y}}_0, \mathbf{\hat{y}}_2)$ to approximate the true posterior distribution. Specifically, we have:
	\begin{align} \label{distortion_2tart}
		&I(\mathbf{x}_0;\mathbf{\hat{y}}_0,\mathbf{\hat{y}}_2) \notag \\
		&= \int p(\mathbf{x}_0, \mathbf{\hat{y}}_0, \mathbf{\hat{y}}_2) \log \frac{p(\mathbf{x}_0|\mathbf{\hat{y}}_0, \mathbf{\hat{y}}_2)}{p(\mathbf{x}_0)} \,d\mathbf{x}_0\,d\mathbf{\hat{y}}_0 \,d\mathbf{\hat{y}}_2 \\
		&\geq \int p(\mathbf{x}_0, \mathbf{\hat{y}}_0, \mathbf{\hat{y}}_2) \log q(\mathbf{x}_0|\mathbf{\hat{y}}_0, \mathbf{\hat{y}}_2) \,d\mathbf{x}_0\,d\mathbf{\hat{y}}_0 \,d\mathbf{\hat{y}}_2 + H(\mathbf{x}_0), \notag
	\end{align}
	where $H(\mathbf{x}_0)$ is the entropy of the source $\mathbf{x}_0$,  which depends solely on the dataset and can thus be ignored during optimization. In this work, we model $q(\mathbf{x}_0|\mathbf{\hat{y}}_0, \mathbf{\hat{y}}_2)$ as a Gaussian distribution, which is given by:
	\begin{equation}
		q(\mathbf{x}_0|\mathbf{\hat{y}}_0, \mathbf{\hat{y}}_2) = \mathcal{N}(\mathbf{x}_0|\mathbf{\tilde{x}}_0, (2\lambda_0)^{-1}\mathbf{I}), \mathbf{\tilde{x}}_0 = g_{\bm{\phi_0}}(\mathbf{\hat{y}}_0,\mathbf{\hat{y}}_2),
	\end{equation} 
	thus Eq. (\ref{distortion_2tart}) can be simplified as:
	\begin{equation} \label{distortion_end}
		I(\mathbf{x}_0;\mathbf{\hat{y}}_0,\mathbf{\hat{y}}_2) \geq -\lambda_0 \|\mathbf{x}_0 - \mathbf{\tilde{x}}_0\|^2,
	\end{equation}
	which reduces to the mean-square error (MSE) between $\mathbf{x}_0$ and $\mathbf{\tilde{x}}_0$. Similarly, $I(\mathbf{x}_1;\mathbf{\hat{y}}_1,\mathbf{\hat{y}}_2)$ and $I(\mathbf{t};\mathbf{\hat{y}}_0, \mathbf{\hat{y}}_1,\mathbf{\hat{y}}_2)$ can be estimated using two other Gaussian distributions, resulting in additional MSE terms $\lambda_1 \|\mathbf{x}_1 - \mathbf{\tilde{x}}_1\|^2$ and $\lambda_t \|\mathbf{t} - \mathbf{\tilde{t}}\|^2$, respectively (see Appendix A for details).
	For minimizing $I(\mathbf{x}_0;\mathbf{\hat{y}}_0)$, we introduce a parameterized distribution $q(\mathbf{\hat{y}}_0)$ to approximate the true marginal distribution, thus we have:
	\begin{align} \label{rate_base}
		&I(\mathbf{x}_0;\mathbf{\hat{y}}_0) = \int p(\mathbf{x}_0) p(\mathbf{\hat{y}}_0|\mathbf{x}_0) \log \frac{p(\mathbf{\hat{y}}_0|\mathbf{x}_0)}{p(\mathbf{\hat{y}}_0)} \,d\mathbf{x}_0\,d\mathbf{\hat{y}}_0 \\
		&\leq \int p(\mathbf{x}_0) p(\mathbf{\hat{y}}_0|\mathbf{x}_0) \big[\log p(\mathbf{\hat{y}}_0|\mathbf{x}_0) - \log q(\mathbf{\hat{y}}_0)\big] \,d\mathbf{x}_0\,d\mathbf{\hat{y}}_0, \notag
	\end{align}
	where the first term can be expanded as:
	\begin{equation}
		p(\mathbf{\hat{y}}_0|\mathbf{x}_0) = \prod_{i} \mathcal{U}(\hat{y}_{0_i}|y_{0_i}-\frac{1}{2},y_{0_i}+\frac{1}{2}), \mathbf{y}_0 = g_{e_0}(\mathbf{x}_0),
	\end{equation}
	where $\mathcal{U}$ denotes a uniform distribution centered at $y_{0_i}$, as a substitute for quantization to enable training, and $g_{e_0}$ is the image-specific pre-encoder. Since the width of the uniform distribution is fixed (equal to one), the first term can be  ignored during optimization. As the latent $\mathbf{\hat{y}}_0$ is discrete, we could regard the second term as the rate. 
	Similarly, $I(\mathbf{x}_0, \mathbf{x}_1; \mathbf{\hat{y}}_2)$ and $I(\mathbf{x}_1;\mathbf{\hat{y}}_1)$ are estimated using two additional parameterized distributions, yielding two more rate terms (see Appendix A for details).
	
	\subsubsection{Enhanced Bayesian Model Optimization} \label{enhance}
	\begin{figure}[t]
		\begin{centering}
			\includegraphics[width=0.5 \textwidth]{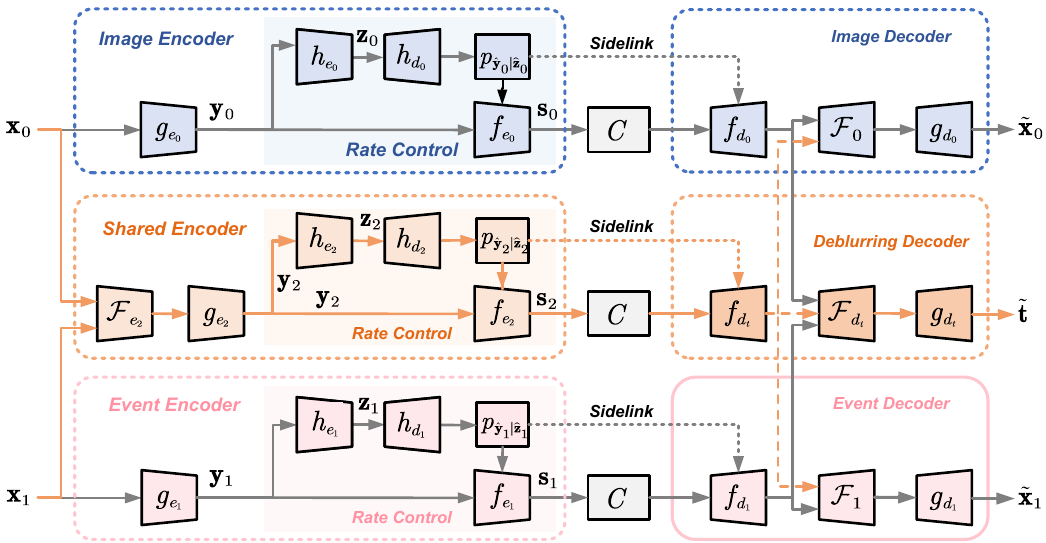}
			\caption{Operational diagram of EBM-based transmission scheme.}
			\label{diagram}
		\end{centering}
	\end{figure}
	
	The base model exhibits certain limitations, as its simplistic non-parametric approach may fail to fully capture the spatial dependencies inherent in each extracted feature.
	To address this issue, we incorporate three hyperprior variables, $\mathbf{\hat{z}}_0, \mathbf{\hat{z}}_1$ and  $\mathbf{\hat{z}}_2$  to construct Enhanced Bayesian Models (EBMs), as illustrated in Fig. \ref{Bayesian2}. These variables effectively model the spatial dependencies within $\mathbf{\hat{y}}_0, \mathbf{\hat{y}}_1$, and $\mathbf{\hat{y}}_2$, respectively. By incorporating the graphical structures shown in Fig. \ref{Bayesian2} into Eq. \eqref{original}, we derive the empirical objective as follows:
	\begin{align}
		&\mathop{\min}\limits_{\mathbf{\hat{y}}_0, \mathbf{\hat{y}}_1, \mathbf{\hat{y}}_2, \mathbf{\hat{z}}_0, \mathbf{\hat{z}}_1, \mathbf{\hat{z}}_2}L_{DisTIB}= \\
		&-\big[I(\mathbf{x}_0; \mathbf{\hat{y}}_0, \mathbf{\hat{y}}_2)
		+ I(\mathbf{x}_1; \mathbf{\hat{y}}_1, \mathbf{\hat{y}}_2) + I(\mathbf{t}; \mathbf{\hat{y}}_0, \mathbf{\hat{y}}_2, \mathbf{\hat{y}}_1) \notag \\
		&+ I(\mathbf{\hat{y}}_0; \mathbf{\hat{z}}_0) + I(\mathbf{\hat{y}}_1; \mathbf{\hat{z}}_1) + I(\mathbf{\hat{y}}_2; \mathbf{\hat{z}}_2) \big] \notag \\
		&+ \beta \big[I(\mathbf{x}_0; \mathbf{\hat{y}}_0)+ I(\mathbf{x}_0, \mathbf{x}_1; \mathbf{\hat{y}}_2) + I(\mathbf{x}_1; \mathbf{\hat{y}}_1) + I(\mathbf{t}; \mathbf{\tilde{x}}_0, \mathbf{\tilde{x}}_1) \notag \\
		&+ I(\mathbf{\hat{z}}_0; \mathbf{x}_0) + I(\mathbf{\hat{z}}_1; \mathbf{x}_1) + I(\mathbf{\hat{z}}_2; \mathbf{x}_0, \mathbf{x}_1)\big]. \notag
	\end{align}
	\begin{figure*}[t]
		\begin{centering}
			\includegraphics[width=0.9 \textwidth]{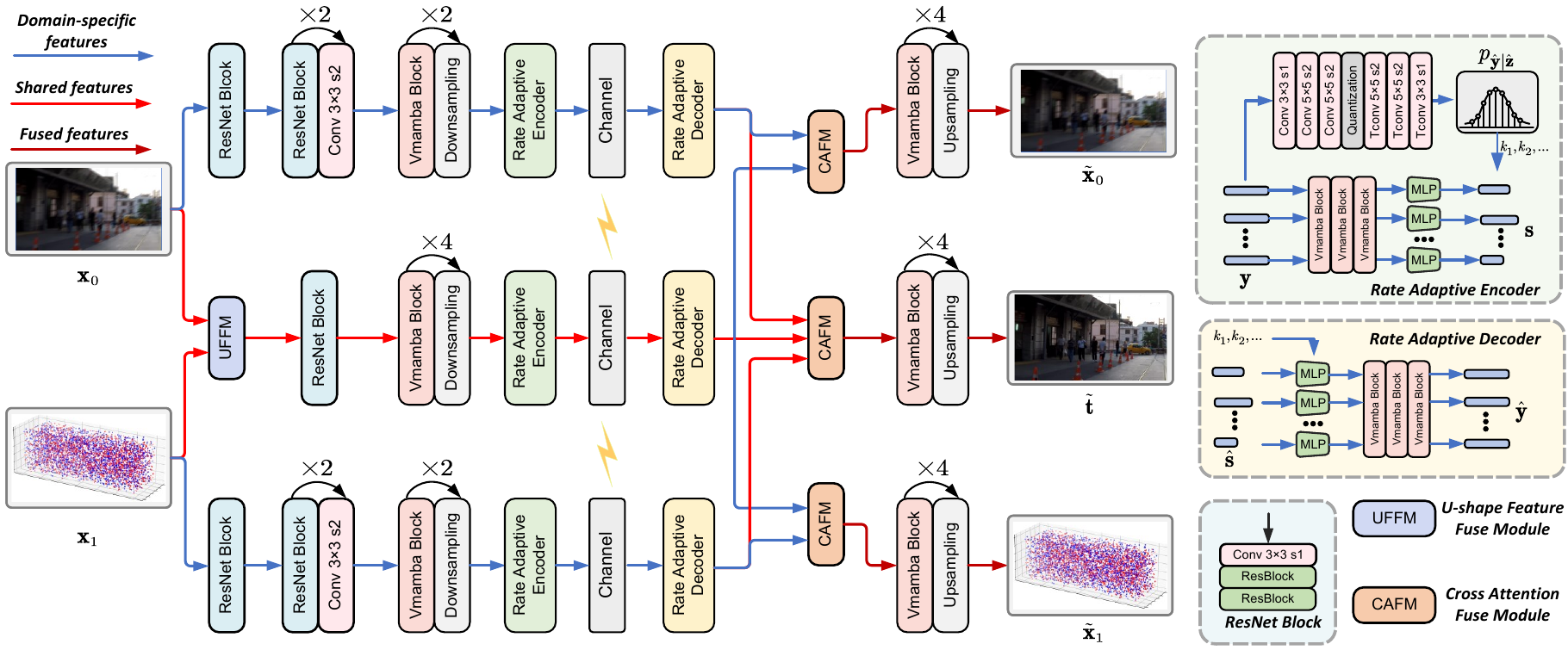}
			\caption{The architecture of the proposed JEIT. The shared and domain-specific features are disentangled at the transmitter, respectively, and fused at the receiver for reconstruction or deblurring.}
			\label{architecture}
		\end{centering}
	\end{figure*}
	
	Similar to the derivation from Eq. \eqref{distortion_2tart} to Eq. \eqref{distortion_end}, the first three terms can be estimated using three Gaussian distributions, leading to three MSE terms: $\lambda_0 \|\mathbf{x}_0 - \mathbf{\tilde{x}}_0\|^2$, $\lambda_1 \|\mathbf{x}_1 - \mathbf{\tilde{x}}_1\|^2$, and $\lambda_t \|\mathbf{t} - \mathbf{\tilde{t}}\|^2$.
	
	To minimize $I(\mathbf{\hat{z}}_0;\mathbf{x}_0)$, similar to Eq. \eqref{rate_base}, we have:
	\begin{align}
		&I(\mathbf{x}_0;\mathbf{\hat{z}}_0) = \int p(\mathbf{x}_0) p(\mathbf{\hat{z}}_0|\mathbf{x}_0) \log \frac{p(\mathbf{\hat{z}}_0|\mathbf{x}_0)}{p(\mathbf{\hat{z}}_0)} \,d\mathbf{x}_0\,d\mathbf{\hat{z}}_0 \\
		&\leq \int p(\mathbf{x}_0) p(\mathbf{\hat{z}}_0|\mathbf{x}_0) \big[\log p(\mathbf{\hat{z}}_0|\mathbf{x}_0) - \log q(\mathbf{\hat{z}}_0)\big] \,d\mathbf{x}_0\,d\mathbf{\hat{z}}_0. \notag
	\end{align}
	Here, the first term can be expressed as:
	\begin{equation} \label{uniform}
		p(\mathbf{\hat{z}}_0|\mathbf{x}_0) = \prod_{i} \mathcal{U}(\hat{z}_{0_i}|z_{0_i}-\frac{1}{2},z_{0_i}+\frac{1}{2}), \mathbf{z}_0 = h_{e_0}(\mathbf{y}_0),
	\end{equation}
	where $h_{e_0}$ represents the hyperprior encoder. The first term in the integral can be omitted from the loss function, while the second term corresponds to the rate, reflecting the bit cost of encoding $\mathbf{\hat{z}}_0.$ Specifically, we model $q(\mathbf{\hat{z}}_0)$ as a non-parametric, fully factorized density, given by:
	
	\begin{equation}
		q(\mathbf{\hat{z}}_0|\bm{\psi}_0) = \prod_{i}\left(q_{z_{0_i}|\bm{\psi}_0^{(i)}}(\bm{\psi}_0^{(i)})*\mathcal{U}(-\frac{1}{2}, \frac{1}{2})\right)(\hat{z}_{0_i}),
	\end{equation}
	
	where $\bm{\psi}_0$ denotes the trainable parameters. We implement non-parametric model using the cumulative distribution function method \cite{balle_arxiv2018}. Similarly, $I(\mathbf{x}_1;\mathbf{\hat{z}}_1)$ and $I(\mathbf{\hat{z}}_2; \mathbf{x}_0, \mathbf{x}_1)$ can be formulated in the same manner, representing the rate terms for $\mathbf{\hat{z}}_1$ and $\mathbf{\hat{z}}_2$, respectively (see Appendix B for details).
	
	Subsequently, we jointly optimize $\big[-I(\mathbf{\hat{y}}_0; \mathbf{\hat{z}}_0) + \beta I(\mathbf{\hat{y}}_0; \mathbf{x}_0)\big]$. When $\beta=1$, this expression becomes
	\begin{align} \label{rate_hyper_2tart}
		&-I(\mathbf{\hat{y}}_0; \mathbf{\hat{z}}_0) + I(\mathbf{\hat{y}}_0; \mathbf{x}_0) \!=\! -\!\!\int\!\!\! p(\mathbf{\hat{y}}_0, \mathbf{\hat{z}}_0) \log \frac{p(\mathbf{\hat{y}}_0|\mathbf{\hat{z}}_0)}{p(\mathbf{\hat{y}}_0)} \,d\mathbf{\hat{y}}_0\,d\mathbf{\hat{z}}_0 \notag \\
		&+ \int p(\mathbf{\hat{y}}_0, \mathbf{x}_0) \log \frac{p(\mathbf{\hat{y}}_0|\mathbf{x}_0)}{p(\mathbf{\hat{y}}_0)} \,d\mathbf{x}_0\,d\mathbf{\hat{y}}_0 \\
		&\leq\!\!\! \int\!\! p(\mathbf{x}_0, \mathbf{\hat{y}}_0, \mathbf{\hat{z}}_0) \big[-\log q(\mathbf{\hat{y}}_0|\mathbf{\hat{z}}_0) \!+\! \log p(\mathbf{\hat{y}}_0|\mathbf{x}_0)\big] \,d\mathbf{x}_0\,d\mathbf{\hat{y}}_0\,d\mathbf{\hat{z}}_0. \notag
	\end{align}
	Here, we utilize the variational distribution $q(\mathbf{\hat{y}}_0|\mathbf{\hat{z}}_0)$ to approximate the true posterior $p(\mathbf{\hat{y}}_0|\mathbf{\hat{z}}_0)$. Similar to Eq. \eqref{uniform}, the second term corresponds to a uniform distribution, and thus can be omitted from the loss function. For the first term, we model it as a Gaussian distribution convolved with a standard uniform distribution:
	\begin{equation} \label{rate_hyper_end}
		\begin{aligned}
			q(\mathbf{\hat{y}}_0|\mathbf{\hat{z}}_0) = \prod_{i} \left(\mathcal{N}(\hat{\mu}_{0_i},\hat{\sigma}_{0_i}^2) *\mathcal{U}(-\frac{1}{2}, \frac{1}{2})\right)(\hat{y}_{0_i}),
		\end{aligned}
	\end{equation}
	where $(\bm{\hat{\mu}}_0, \bm{\hat{\sigma}}_0) = h_{d_0}(\mathbf{\hat{z}}_0)$, with $\bm{\hat{\mu}}_0$ and $\bm{\hat{\sigma}}_0$ representing the predicted mean and standard deviation based on $\mathbf{\hat{z}}_0$, respectively. Therefore, Eq. \eqref{rate_hyper_2tart} effectively reflects the bit length required to encode $\mathbf{\hat{y}}_0$ conditioned on $\mathbf{\hat{z}}_0$. Following this approach, $-I(\mathbf{\hat{y}}_1; \mathbf{\hat{z}}_1) + \beta I(\mathbf{\hat{y}}_1; \mathbf{x}_1)$ and $-I(\mathbf{\hat{y}}_2; \mathbf{\hat{z}}_2) + \beta I(\mathbf{\hat{y}}_2; \mathbf{x}_0, \mathbf{x}_1)$ can be similarly estimated using corresponding variational distributions (see Appendix B for details). These terms correspond to the bit cost for encoding $\mathbf{\hat{y}}_1$ and $\mathbf{\hat{y}}_2$, respectively. Consequently, the overall learning objective for the enhanced model is formulated as:
	\begin{align} \label{loss}
		L &= \mathbb{E}_{\mathbf{x}_0, \mathbf{x}_1, \mathbf{t}} \big[ \lambda_0 \|\mathbf{x}_0 - \mathbf{\tilde{x}}_0\|^2 + \lambda_1 \|\mathbf{x}_1 - \mathbf{\tilde{x}}_1\|^2 + \lambda_t \|\mathbf{t} - \mathbf{\tilde{t}}\|^2 \notag \\
		&\quad - \log q(\mathbf{\hat{y}}_0|\mathbf{\hat{z}}_0) - \log q(\mathbf{\hat{y}}_1|\mathbf{\hat{z}}_1) - \log q(\mathbf{\hat{y}}_2|\mathbf{\hat{z}}_2) \notag \\
		&\quad - \log q(\mathbf{\hat{z}}_0) - \log q(\mathbf{\hat{z}}_1) - \log q(\mathbf{\hat{z}}_2) \big].
	\end{align}
	\begin{figure*}[t]
		\begin{centering}
			\includegraphics[width=0.9 \textwidth]{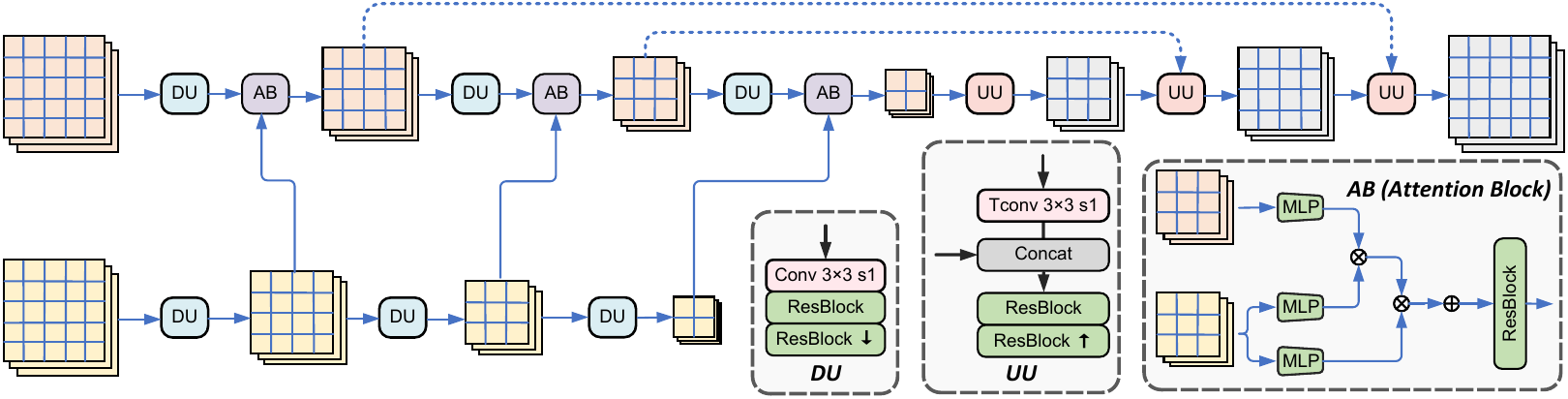}
			\caption{U-shape Feature Fuse Module (UFFM).}
			\label{UFFM}
		\end{centering}
	\end{figure*}
	
	\begin{figure}[t]
		\begin{centering}
			\includegraphics[width=0.46 \textwidth]{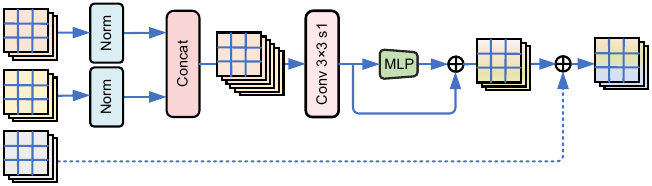}
			\caption{Cross Attention Fuse Module (CAFM). 
				In the domain-specific decoder, solid lines indicate the fusion of domain-specific and shared features, while in the deblurring decoder, solid lines represent E-I data feature fusion, and dotted lines denote the fusion of E-I data and shared features.}
			\label{CAFM}
		\end{centering}
	\end{figure}
	\subsection{EBM-Based Dynamic CBR Allocation for Transmission} \label{CBRallo2}
	As described above, we have successfully disentangled compact and informative shared and domain-specific information. Inspired by \cite{Jincheng_JSAC2022}, we utilize the distributions of the representations $\mathbf{\hat{y}}_0, \mathbf{\hat{y}}_1$, and $\mathbf{\hat{y}}_2$ to indicate the CBR for transmission.
	Specifically, a representation containing more information exhibits higher entropy and therefore requires more symbols for transmission, leading to the adoption of a higher CBR. 
	Typically, $\mathbf{y}$ (denoting $\mathbf{y}_0, \mathbf{y}_1,$ or $\mathbf{y}_2$ for general analysis) is comprised of multiple embedding vectors $y_j$, each with length $C$. After the rate-matching operation, the length of each embedding vector is adjusted to ${k}_j$, where ${k}_j$ is determined based on the prior distribution $p \left(\hat{y}_j | \mathbf{\hat{z}}\right)$. Consequently, optimizing the rate terms in Eq. (\ref{loss}) corresponds to optimizing the CBR allocation. This alignment ensures that our overarching objective in Eq. (\ref{objective}) is effectively realized through the optimization of Eq. (\ref{loss}).
	
	The overall process is outlined in Fig. \ref{diagram}. First, a domain-specific pre-encoder $g_{e_0}/g_{e_1}$ is utilized for each input to extract domain-specific features. Meanwhile, a feature fusion encoder $\mathcal{F}_{e_2}$ is used to deeply extract and fuse the shared features from the E-I data, followed by a shared pre-encoder $g_{e_2}$  that maps the fused features into a low-dimensional representation. Next, a hyperprior encoder $h_{e_i}$ and decoder $h_{d_i}$ are employed to determine the desired length $k_i$ of each symbol vector. A mask encoder $f_{e_i}$ then generates variable-length vectors based on $k_i$. At the receiver, a mask decoder $f_{d_i}$ is first utilized to reconstruct fixed-length vectors. Subsequently, each domain-specific decoder processes its corresponding domain-specific and shared features through a feature fusion module $\mathcal{F}_0/\mathcal{F}_1$, followed by a post-decoder $g_{d_0}/g_{d_1}$ to recover the original source.  In parallel, for the deblurring task, all extracted features are processed through another feature fusion module $\mathcal{F}_{d_t}$, followed by deblurring post decoder $g_{d_t}$ to perform deblurring.

	\section{Model Architecture}\label{SEC4}
	In this section, we  present the detailed description of the model architecture in Section \ref{archi} and \ref{archi2}. Additionally, we introduce an alternative model specifically tailored for deblurring task \ref{case}.

	\subsection{Overall Model Architecture} \label{archi}
	
	The overall architecture of our proposed JEIT is shown in Fig. \ref{architecture}. As stated in the previous section, it comprises an image/event encoder, a shared encoder, an image/event decoder, and a deblurring decoder. Both the encoders and decoders are primarily implemented with multiple ResNet and VMamba blocks. Notably, the rate-adaptive encoders are built upon VMamba  blocks and incorporate a hyperprior model to indicate the mask ratio. Down-sampling is executed through patch merging, while up-sampling reverses this process \cite{vmamba}. The hyperprior model consists of multiple convolutional layers. The mask ratios are also transmitted through wireless channels and serve as critical indicators to aid the rate-adaptive decoders in converting variable-length vectors back into fixed-length vectors.
	
	\begin{figure}[t]
		\begin{centering}
			\includegraphics[width=0.42 \textwidth]{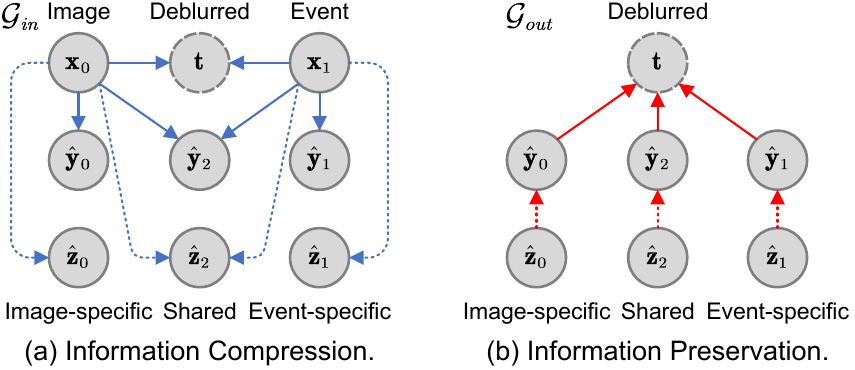}
			\caption{Bayesian networks for JEIT-T.}
			\label{Bayesian3}
		\end{centering}
	\end{figure}
	\subsection{Fusion Module Design} \label{archi2}
	At the shared encoder, extracting shared information from the E-I data is crucial. 
	Traditional feature fusion methods, such as simple concatenation or weighted averaging, often overlook the importance of cross-scale information transfer, leading to the loss of fine spatial details in the fused features. 
	To address this issue, inspired by the U-Net architecture, which enables effective multi-scale feature fusion, we design a U-shape Feature Fuse Module (UFFM), as shown in Fig. \ref{UFFM}. 
	In this module, we first extract hierarchical features from both the event stream and the blurry image using the U-Net encoders, which are composed of multiple down-sampling units (DUs). During the extraction process, extracted features from the blurry image are combined with corresponding event features at each level before being passed to deeper layers. To facilitate effective fusion, we employ an attention block (AB) at each level, ensuring robust interaction between the blurry image and event features.
	The U-Net decoder is constructed using multiple up-sampling units  (UUs) that progressively integrate more deeply fused features. Skip connections (represented by blue dotted lines) directly link each DU output to its corresponding UU, preserving essential spatial details throughout decoding. This approach enables the network to effectively combine complementary information from the E-I data.
	
	\begin{figure*}[t]
		\begin{centering}
			\includegraphics[width=0.95 \textwidth]{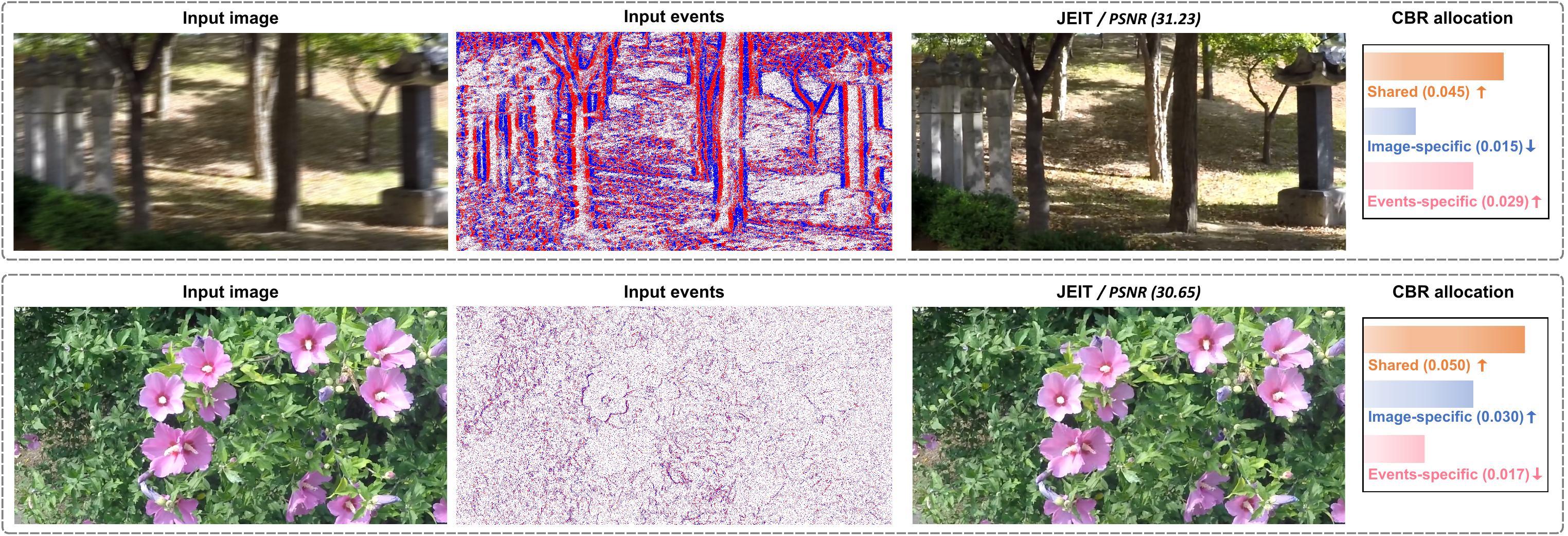}
			\par \end{centering}
		\caption{
			Visualization of rate allocation strategy of shared symbols, image-specific symbols, and events-specific symbols for different images, where two example images are sampled from the GoPro dataset. The upward arrow “↑” indicates “more symbols are allocated”, and vice versa. For each image, we visualize its reconstruction over the AWGN channels at an SNR of $10\text{dB}$.}
		\label{visualization1}
	\end{figure*}
	
	At the receiver, we recover low-dimensional image-specific features, event-specific features, and shared features. For the image and event decoders, our approach involves fusing domain-specific features with shared features. As shown in Fig. \ref{CAFM}, we introduce a Cross Attention Fusion Module (CAFM) to handle this fusion. Features from both branches are first normalized, concatenated, and then processed through a CNN followed by a multi-layer perception (MLP). The final output is obtained by summing the outputs of the CNN and MLP. For the deblurring decoder, which receives three types of features, we first fuse the domain-specific features using the aforementioned method and then integrate the shared features to obtain the final result.

	\subsection{Deblurring Solely Case} \label{case}
	In extreme scenarios where only a deblurred image is prioritized and the reconstruction of E-I data is unnecessary, we adjust JEIT to optimize channel bandwidth efficiency by concentrating solely on generating deblurred images, designating this model as JEIT-T. To realize this adaptation, we redesign the Bayesian networks, as illustrated in Fig. \ref{Bayesian3}. In the information compression stage, we extract task-relevant domain-specific and shared information. During the information preservation stage, optimization is solely directed toward enhancing deblurring performance. Following similar procedures outlined in Section \ref{SEC3}, we derive the loss function as follows:
	\begin{align} \label{loss_propose2}
		L &= \mathbb{E}_{\mathbf{x}_0, \mathbf{x}_1, \mathbf{t}}[\lambda_t \|\mathbf{t} - \mathbf{\tilde{t}}\|^2 \notag \\
		&- \log q(\mathbf{\hat{y}}_0|\mathbf{\hat{z}}_0) - \log q(\mathbf{\hat{y}}_1|\mathbf{\hat{z}}_1) - \log q(\mathbf{\hat{y}}_2|\mathbf{\hat{z}}_2) \notag \\
		& - \log q(\mathbf{\hat{z}}_0) - \log q(\mathbf{\hat{z}}_1) - \log q(\mathbf{\hat{z}}_2)].
	\end{align}
	In terms of architecture design, we streamline the model by omitting the image and event decoders, significantly reducing the model size. Additionally, since only the deblurring image is required at the decoder, we perform feature fusion at the encoder stage. This approach employs a single rate-adaptive encoder to generate the channel input symbols, further minimizing the model size and enhancing computational efficiency.
	
	\begin{figure*}[t]
		\begin{centering}
			\subfloat[]{\label{overhead1}\includegraphics[width=5.5cm]{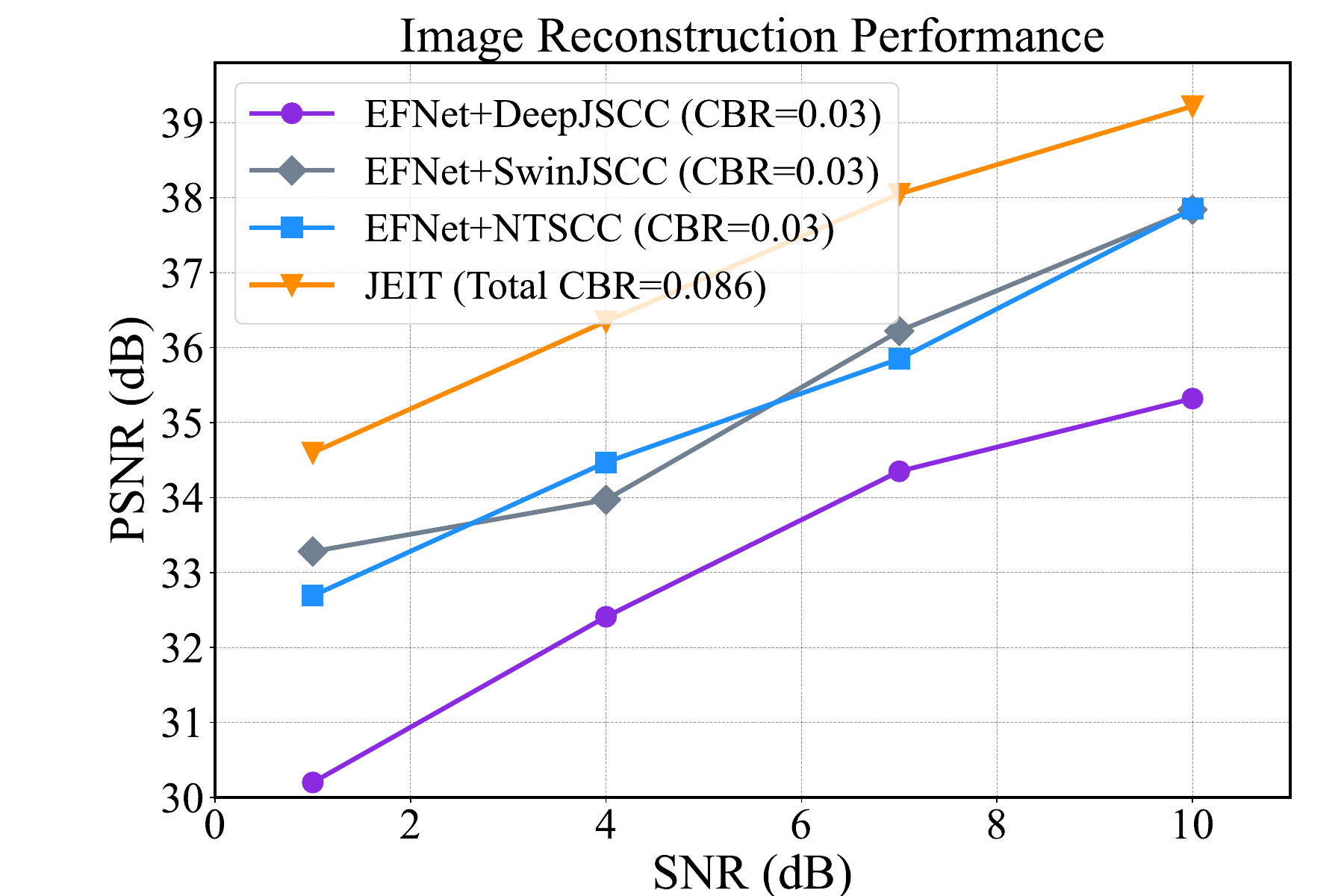}}
			\subfloat[]{\label{overhead1}\includegraphics[width=5.5cm]{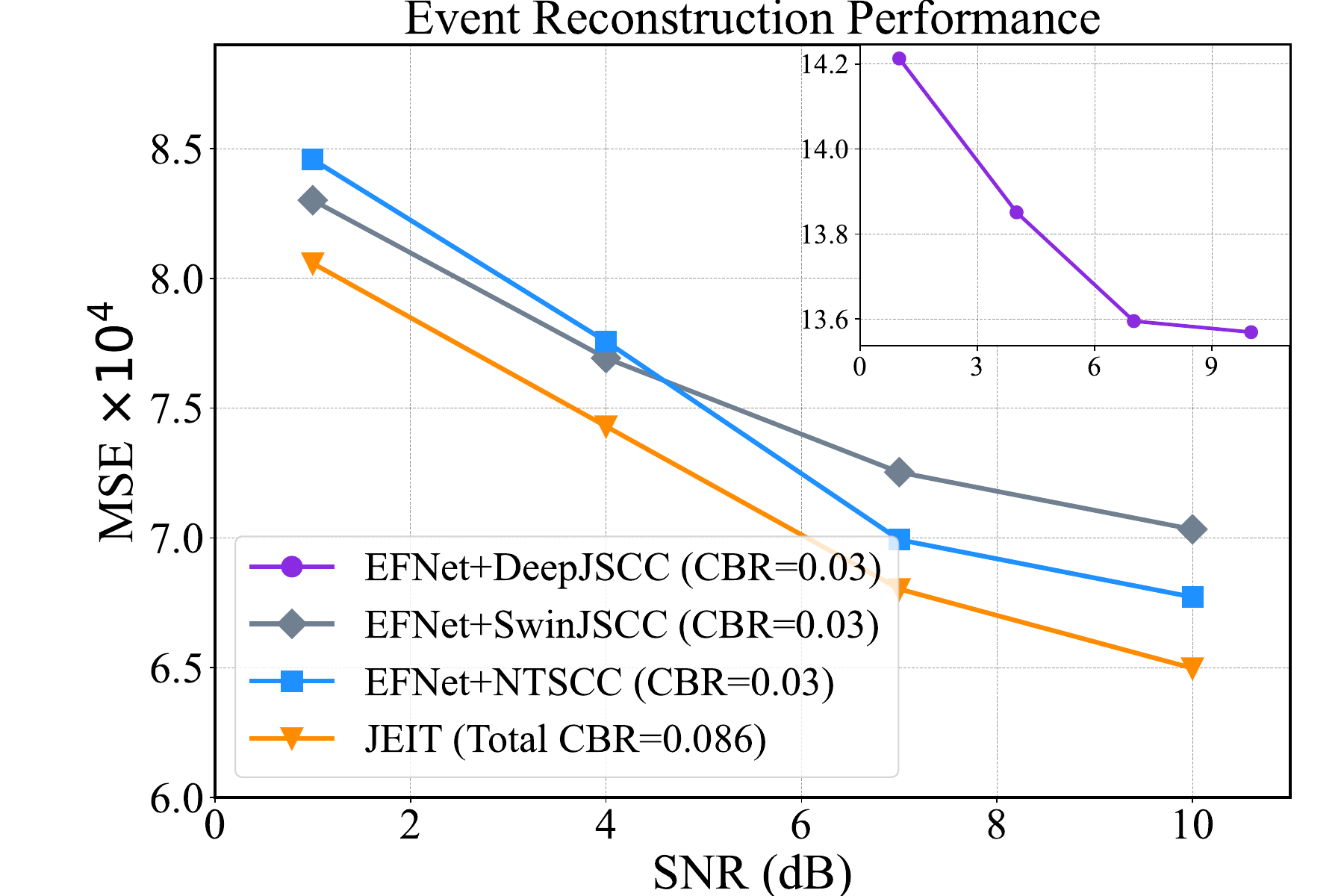}}
			\subfloat[]{\label{overhead1}\includegraphics[width=5.5cm]{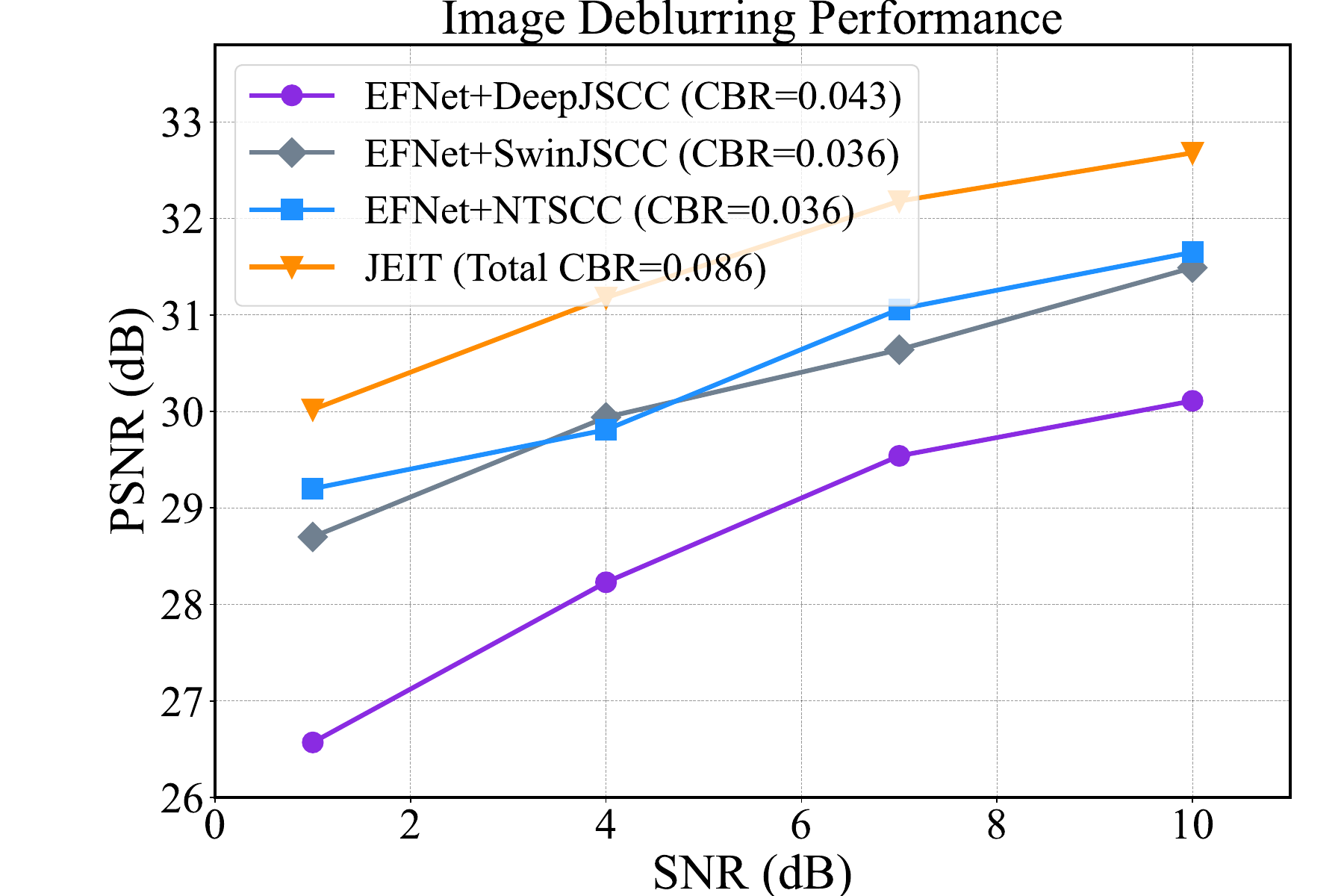}} \\ [-10pt]
			\subfloat[]{\label{overhead1}\includegraphics[width=5.5cm]{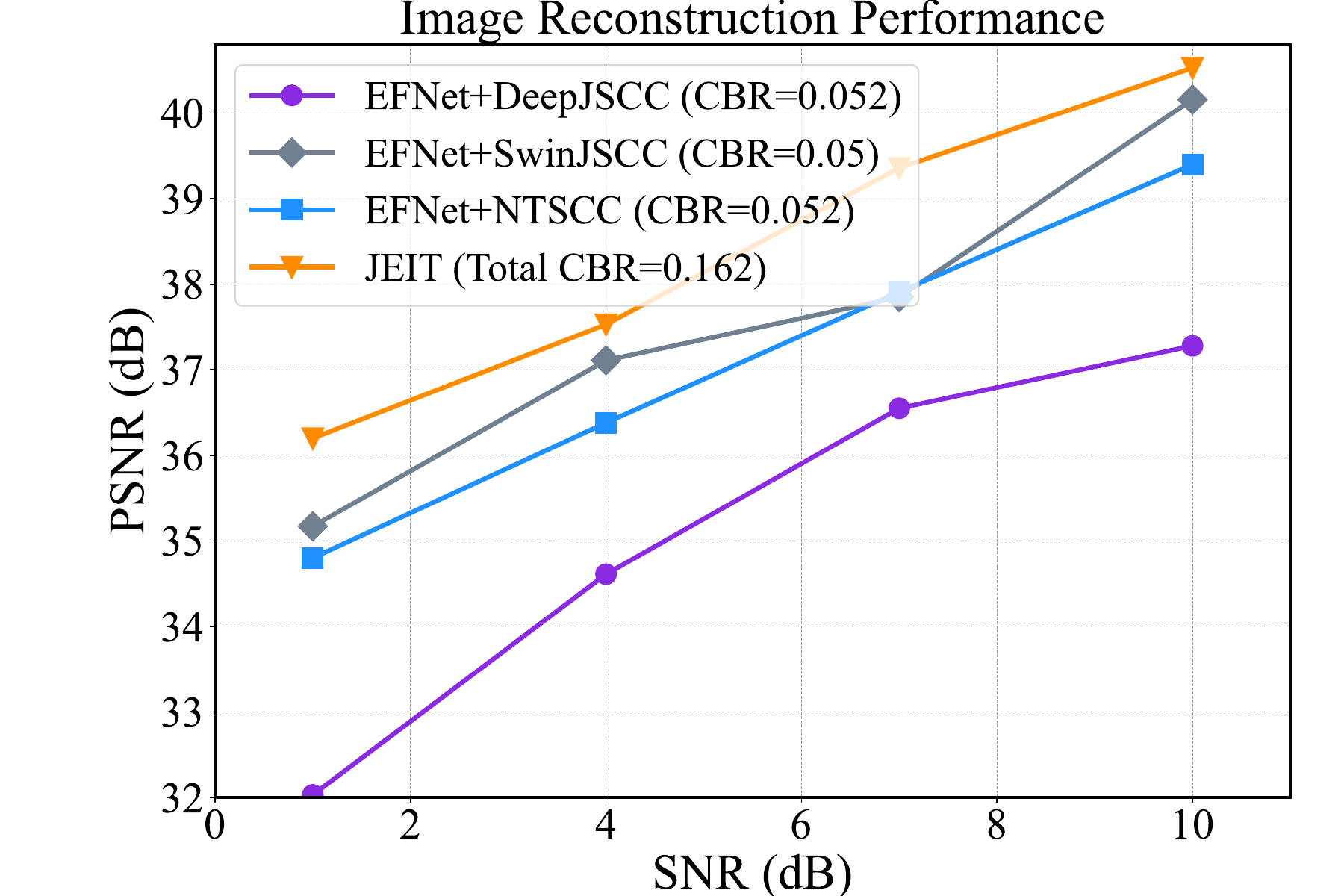}}
			\subfloat[]{\label{overhead1}\includegraphics[width=5.5cm]{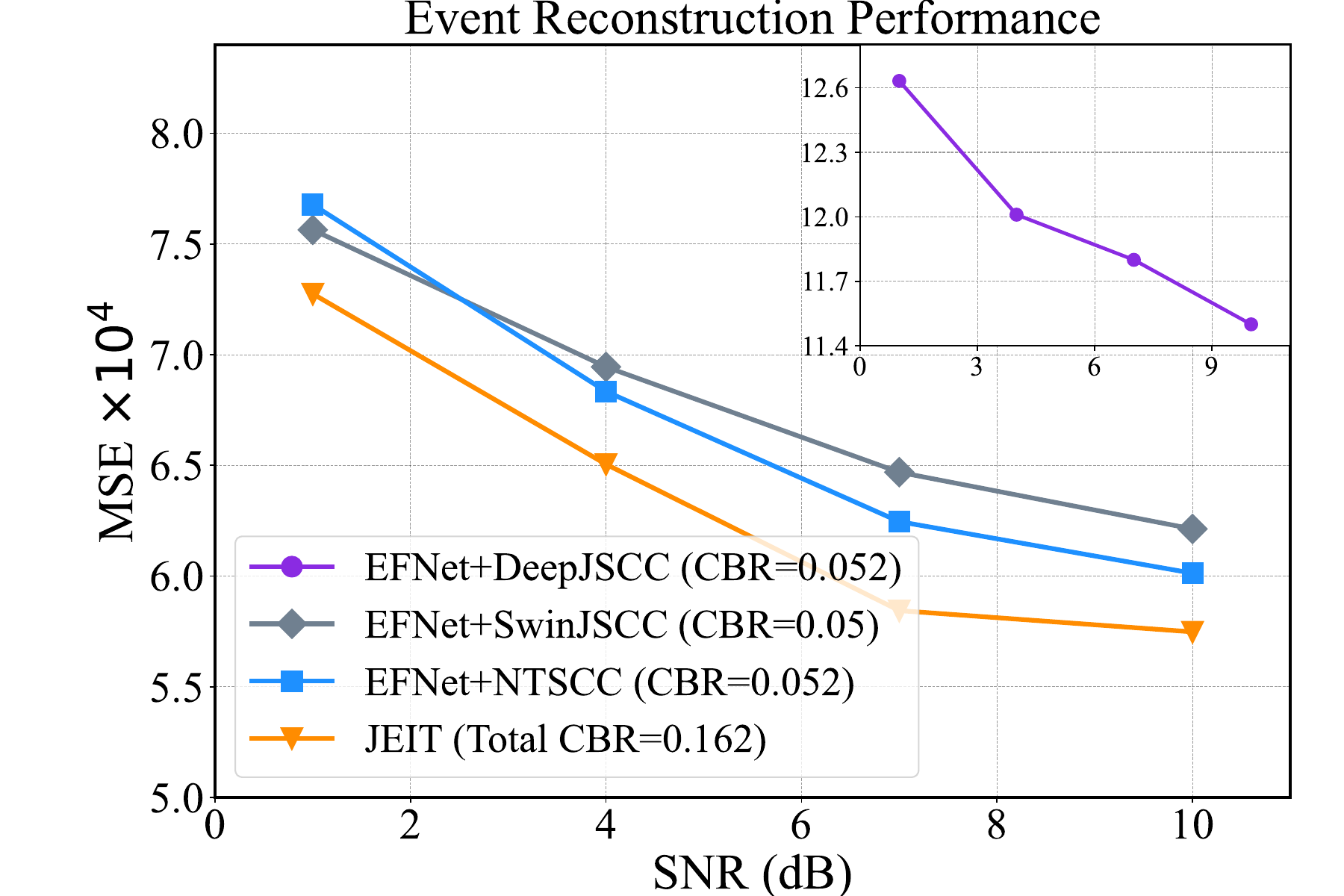}}
			\subfloat[]{\label{overhead1}\includegraphics[width=5.5cm]{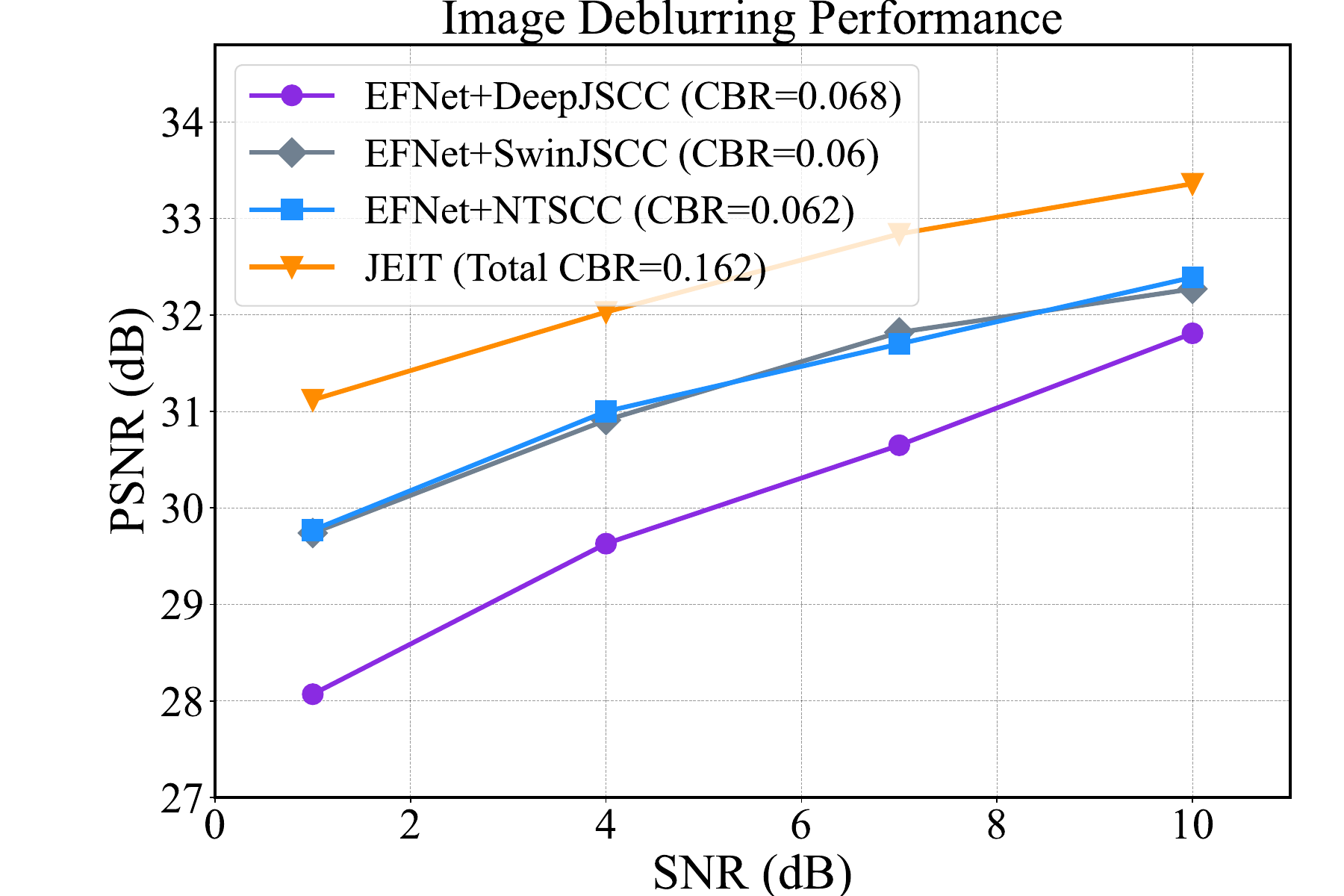}} \\ [-5pt]
			\caption{The performance versus SNR over the AWGN channels for image transmission, events transmission, and deblurring tasks of: (a)$\sim$(c) total $\text{CBR}=0.086$; and (d)$\sim$(f) total $\text{CBR}=0.16$.}
			\label{SNR1}
		\end{centering}
	\end{figure*}
	\section{Experiments}\label{SEC5}
	\subsection{Setup}
	\subsubsection{Datasets}
	We initially train and evaluate our scheme on the GoPro dataset \cite{gopro}, which consists of $3,214$ pairs of blurry images and clear images, with a resolution of $1,280\times720$. The dataset is divided into $2,103$ training images and $1,111$ testing images. Specifically,  the blurry images are generated by averaging multiple high-speed clear images. To simulate events, we resort to ESIM \cite{esim}, an open-source event camera simulator, to generate corresponding event streams. For training, the images are randomly cropped to a size of $256 \times 256$. 
	
	Additionally, we evaluate our method on the REBlur dataset \cite{sun2022event}, which consists of $1,469$ groups of $260\times360$ blurry-sharp image pairs with associated events, where $486$ pairs are used for training and $983$ pairs for testing. In REBlur, blurry images and events are captured using a pair of time-synchronized and spatially-aligned RGB and event cameras during the initial stage. Sharp images are later captured by relocating the slide rail to match the position during the midpoint of the blurry image’s exposure. For both training and evaluation on REBlur, images are randomly cropped to $256 \times 256$. 
	
	\begin{figure*}[!htbp]
		\begin{centering}
			\subfloat[on the GoPro Dataset]{\label{overhead1}\includegraphics[width=4.5cm]{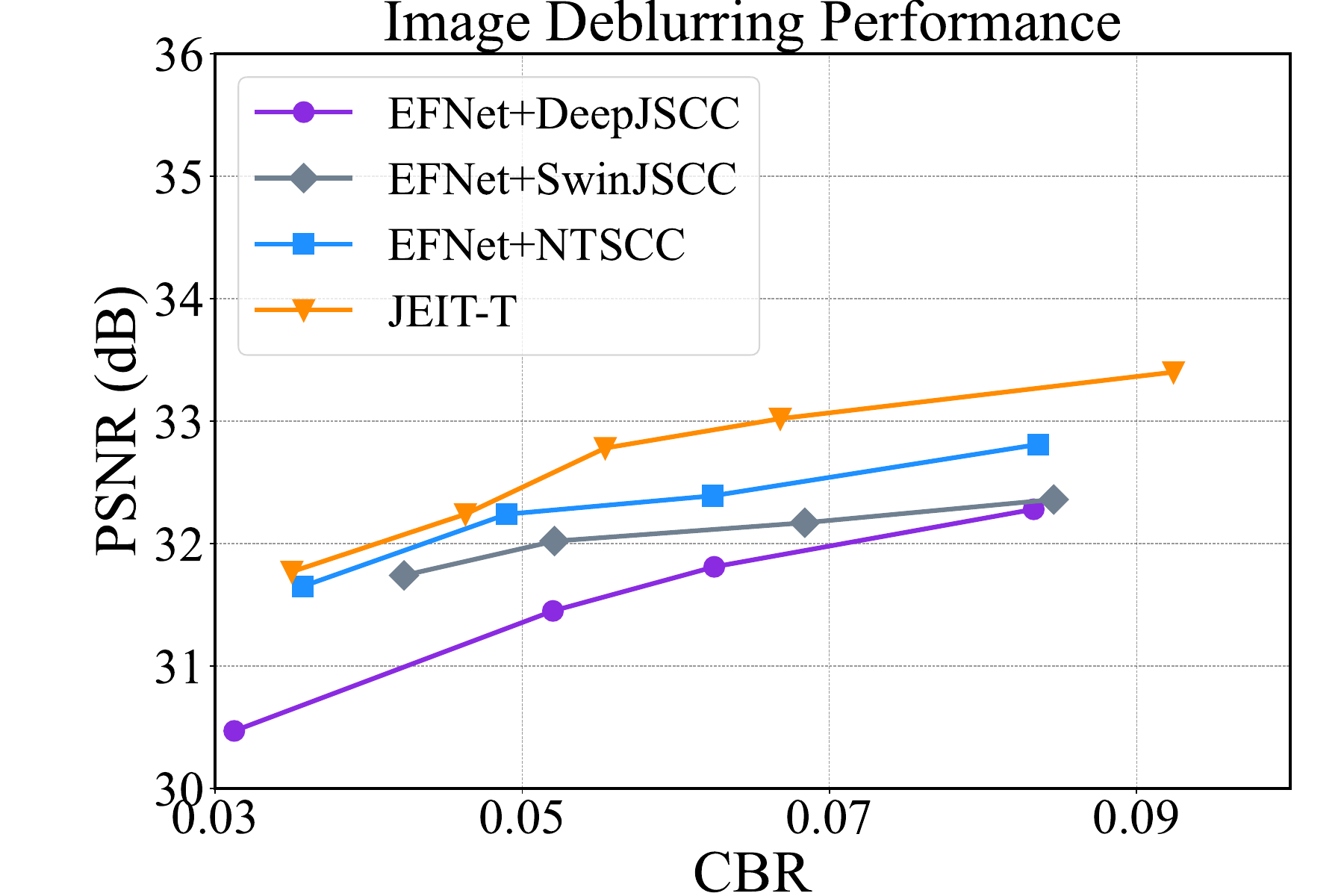}}
			\subfloat[on the GoPro Dataset]{\label{overhead1}\includegraphics[width=4.5cm]{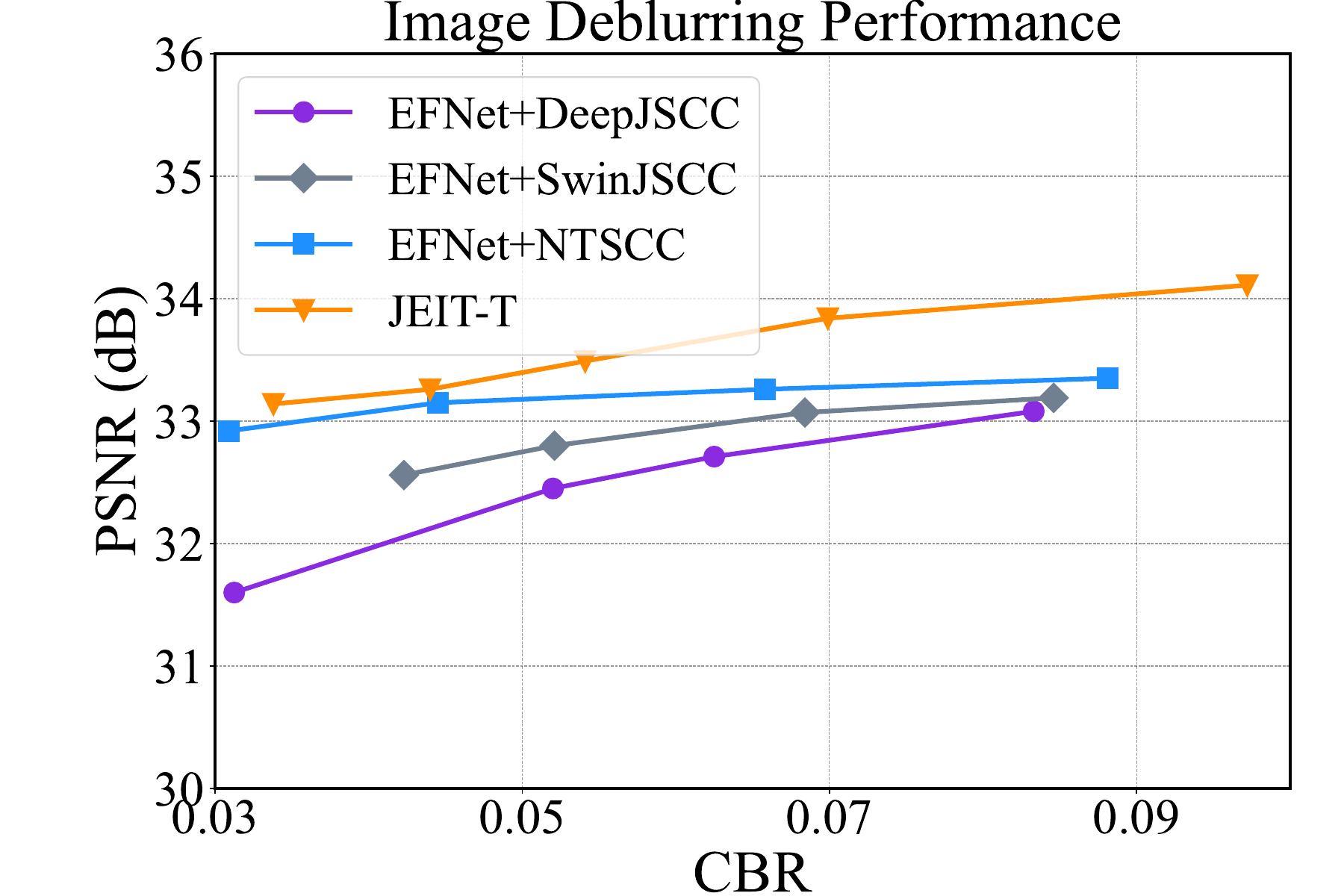}}
			\subfloat[on the REBlur Dataset]{\label{overhead1}\includegraphics[width=4.5cm]{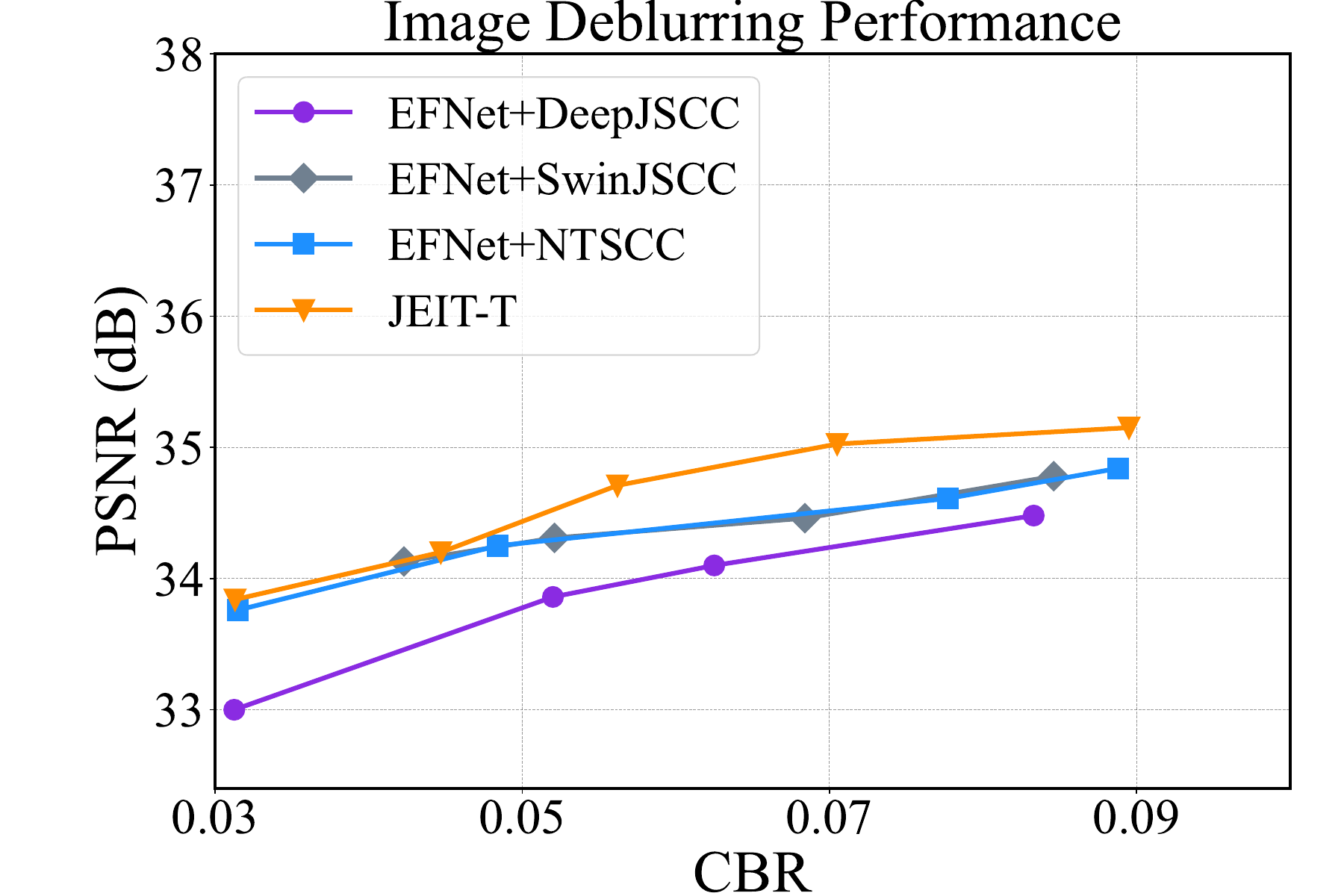}}
			\subfloat[on the REBlur Dataset]{\label{overhead1}\includegraphics[width=4.5cm]{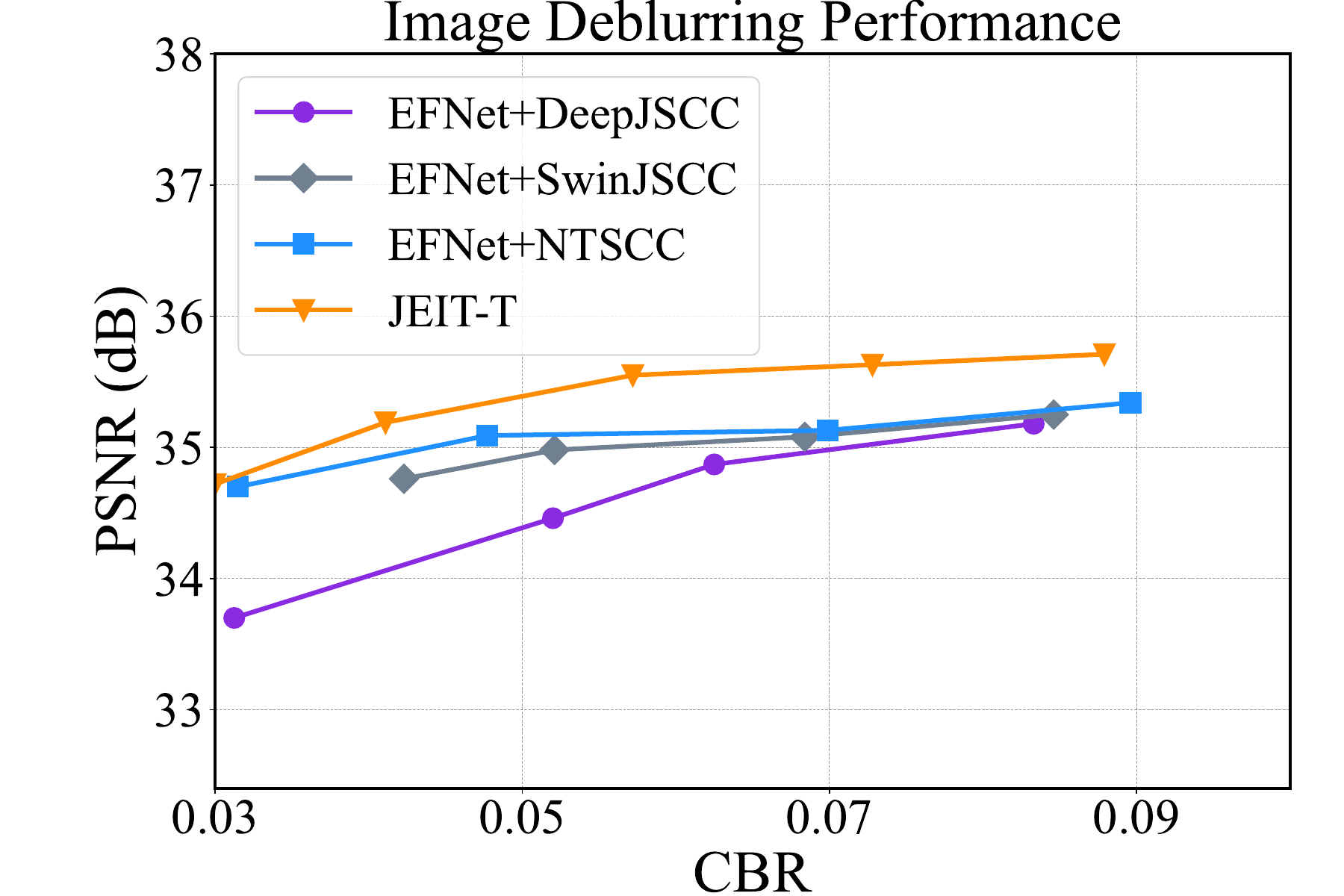}}
			\caption{The deblurring performance versus CBR at (a) $\text{SNR}=10 \text{dB}$ on the GoPro dataset, (b) $\text{SNR}=18 \text{dB}$ on the GoPro dataset, (c) $\text{SNR}=10 \text{dB}$ on the REBlur dataset, and (d) $\text{SNR}=18 \text{dB}$ on the REBlur dataset.}
			\label{CBR1}
		\end{centering}
	\end{figure*}
	
	\subsubsection{Implementation Details}
	We implement both the benchmarks and our proposed methods using PyTorch. We utilize the Adam optimizer with a batch size of $3$ for JEIT, and a batch size of $6$ for JEIT-T. The initial learning rate is set to $1\times 10^{-4}$ and is scheduled to decay after several epochs. The models are trained with $\eta=0.24$. Moreover, we investigate the performance of the proposed methods under the widely-used AWGN channels.
	
	To assess image quality,  we adopt the peak signal-to-noise ratio (PSNR) as the evaluation metric, which measures the distortion between the ground-truth image and the received image.
	For events, we set $M=3$ for the unified event representation and normalize the tensors by the maximum value to the $[-1, 1]$ range. 
	As there is no standardized metric for event evaluation, we measure event distortion by calculating the MSE between the transmitted and received events.

	\subsubsection{Benchmarks}
	To comprehensively evaluate the effectiveness of the proposed methods, we incorporate several state-of-the-art mechanisms for comparison. 
	First, we integrate the emerging image transmission system DeepJSCC to transmit the initial blurry images. For event transmission, we adapt DeepJSCC by modifying the number of input channels to $6$. Regarding image deblurring, we establish an integrated pipeline by combining DeepJSCC with the advanced deblurring network EFNet \cite{sun2022event}, employing a joint training strategy to optimize end-to-end performance. Specifically, events and the blurry image are processed through EFNet at the transmitter to generate a deblurred image, which is then transmitted via DeepJSCC to reconstruct the deblurred image at the receiver. This joint optimization approach ensures maximum overall performance and provides a fair comparison with our proposed framework; without it, performance would be significantly compromised. In particular, we adopt the enhanced DeepJSCC scheme trained with a single signal-to-noise ratio (SNR) from \cite{ADJSCC}, which has been shown to outperform the original implementation in \cite{Eirina_TCCN2019}. We refer to this method as `EFNet+DeepJSCC'.
	In addition to DeepJSCC, we include NTSCC \cite{Jincheng_JSAC2022} as a benchmark to highlight the relative performance gains of our approach. Furthermore, we incorporate SwinJSCC \cite{SwinJSCC}, an advanced image transmission scheme based on the Swin-transformer architecture. To ensure consistency across all methods, we apply the same modifications to NTSCC and SwinJSCC as we do to DeepJSCC, enabling them to effectively transmit event data and perform deblurring task. These adapted versions are referred to as `EFNet+NTSCC' and `EFNet+SwinJSCC', respectively. This comprehensive benchmarking framework allows for a rigorous and fair demonstration of the advantages offered by our proposed methods.
	\begin{figure*}[t]
		\begin{centering}
			\includegraphics[width=0.75 \textwidth]{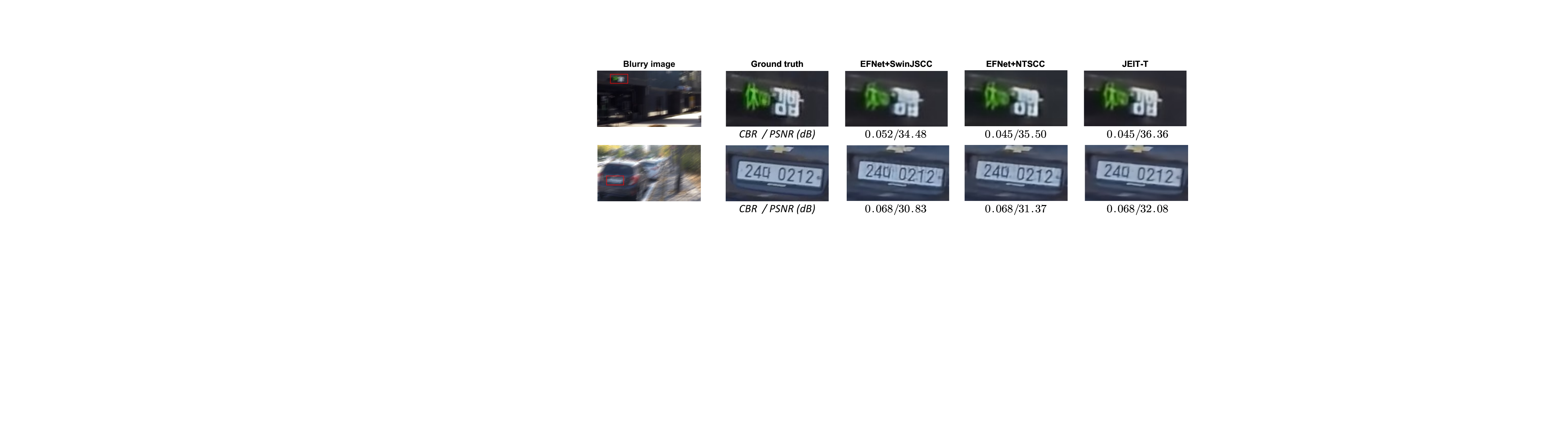}
			\par \end{centering}
		\caption{Visualization example of the deblurred images, where the metrics are [CBR/PSNR]. These reconstructed images are obtained by employing different schemes over the AWGN channels at the SNR of $10 \text{dB}$.}
		\label{visualization2}
	\end{figure*}	
	
	\subsection{Visualization of Dynamic CBR Allocation}
	We first analyze  the symbol allocation results of the model trained on the GoPro dataset, as illustrated in Fig. \ref{visualization1}. A notable difference in symbol lengths is observed between the two selected images. For the upper image, which exhibits severe blurring, capturing more motion information is crucial. Consequently, a larger proportion of symbols is allocated to events, which effectively capture dynamic scene changes. In contrast, the lower image is relatively static with richer color information. Given the minimal motion, the event data carries less valuable information, leading to fewer symbols being assigned to event transmission. This figure shows that our proposed scheme realizes dynamic bandwidth allocation in different scenes with different dynamics.
	\subsection{Performance Comparison with Benchmarks}
	In Fig. \ref{SNR1}, we present a comparison of the PSNR and MSE results obtained using different methods on the GoPro dataset across varying SNR levels under AWGN channels. To ensure a fair comparison, we maintain a similar average total CBR for the three tasks across all methods.
	For instance, the `EFNet + NTSCC' method allocates CBRs of $0.03, 0.03$, and $0.036$ for image reconstruction, event reconstruction, and image deblurring, respectively, yielding a total CBR of $0.086$. Similarly, our proposed JEIT framework achieves simultaneous reconstruction and deblurring tasks with a total CBR of approximately $0.086$, matching the bandwidth consumption of the baseline methods.
	The results demonstrate that our proposed joint transmission and deblurring scheme consistently outperforms all other methods across all tasks. Notably, the performance gap between `EFNet + DeepJSCC' and JEIT is  substantial,  with JEIT achieving approximately $17\% $ CBR savings, highlighting its superior efficiency.
	Additionally, Figs. \ref{SNR1}(d) to \ref{SNR1}(f) illustrate the performance comparison at a higher total CBR of $0.16$. The trends observed at a CBR of $0.16$ are consistent with those at $0.086$, demonstrating the strong generalization capability of JEIT.
	Moreover, by comparing Figs. \ref{SNR1}(a) to \ref{SNR1}(c) and Figs. \ref{SNR1}(d) to \ref{SNR1}(f), the performance gap between our proposed model and the benchmarks  becomes even more pronounced at lower CBRs. This is primarily because, under low CBR conditions, separately transmitting features for each task leads to extremely limited bandwidth allocation per task, which often fails to capture sufficient effective features. In contrast, JEIT simultaneously captures both shared and domain-specific information, eliminating redundancy and ensuring the efficient transmission of critical features.
	
	\begin{figure*}[t]
		\begin{centering}
			\subfloat[]{\label{overhead1}\includegraphics[width=5.5cm]{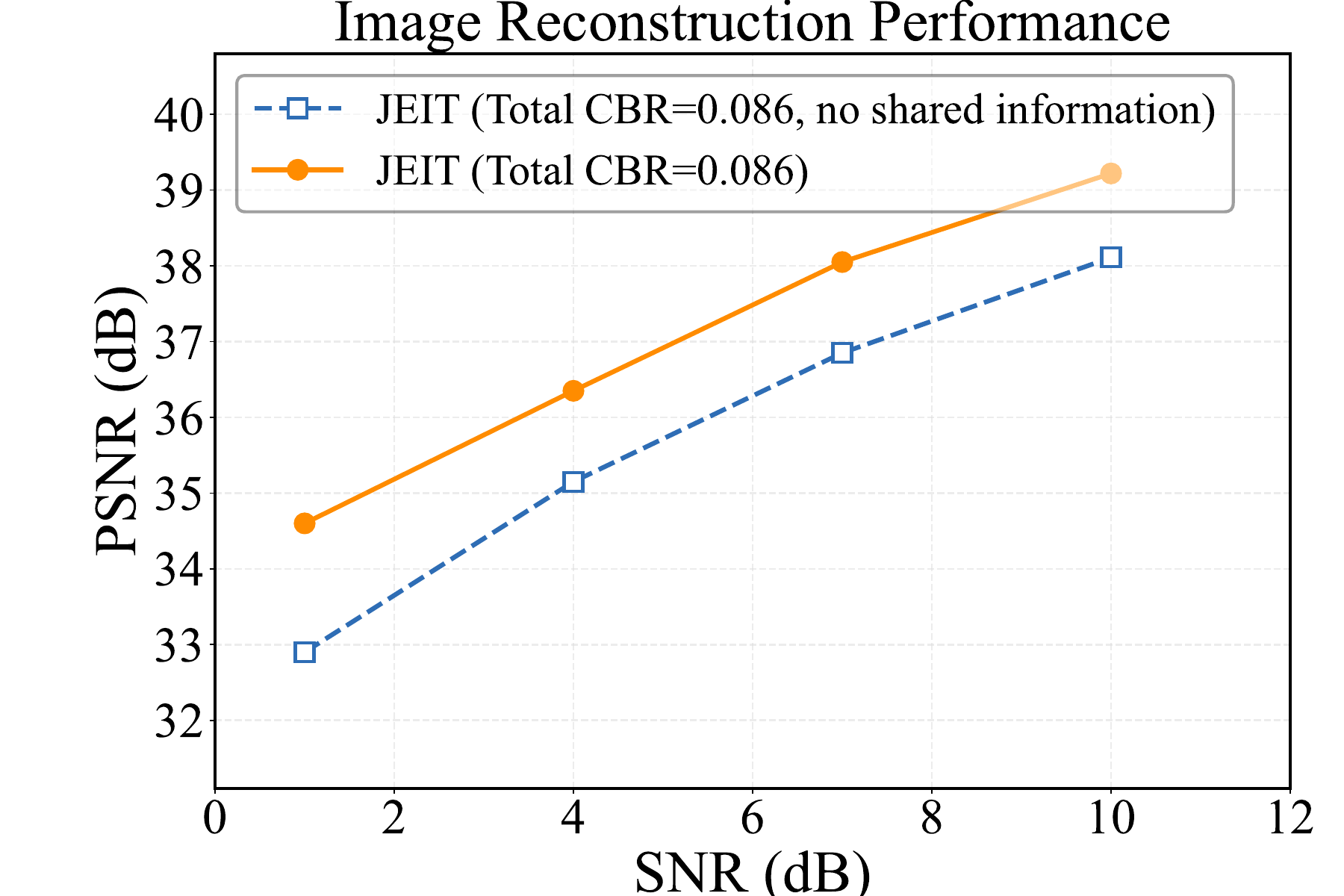}}
			\subfloat[]{\label{overhead1}\includegraphics[width=5.5cm]{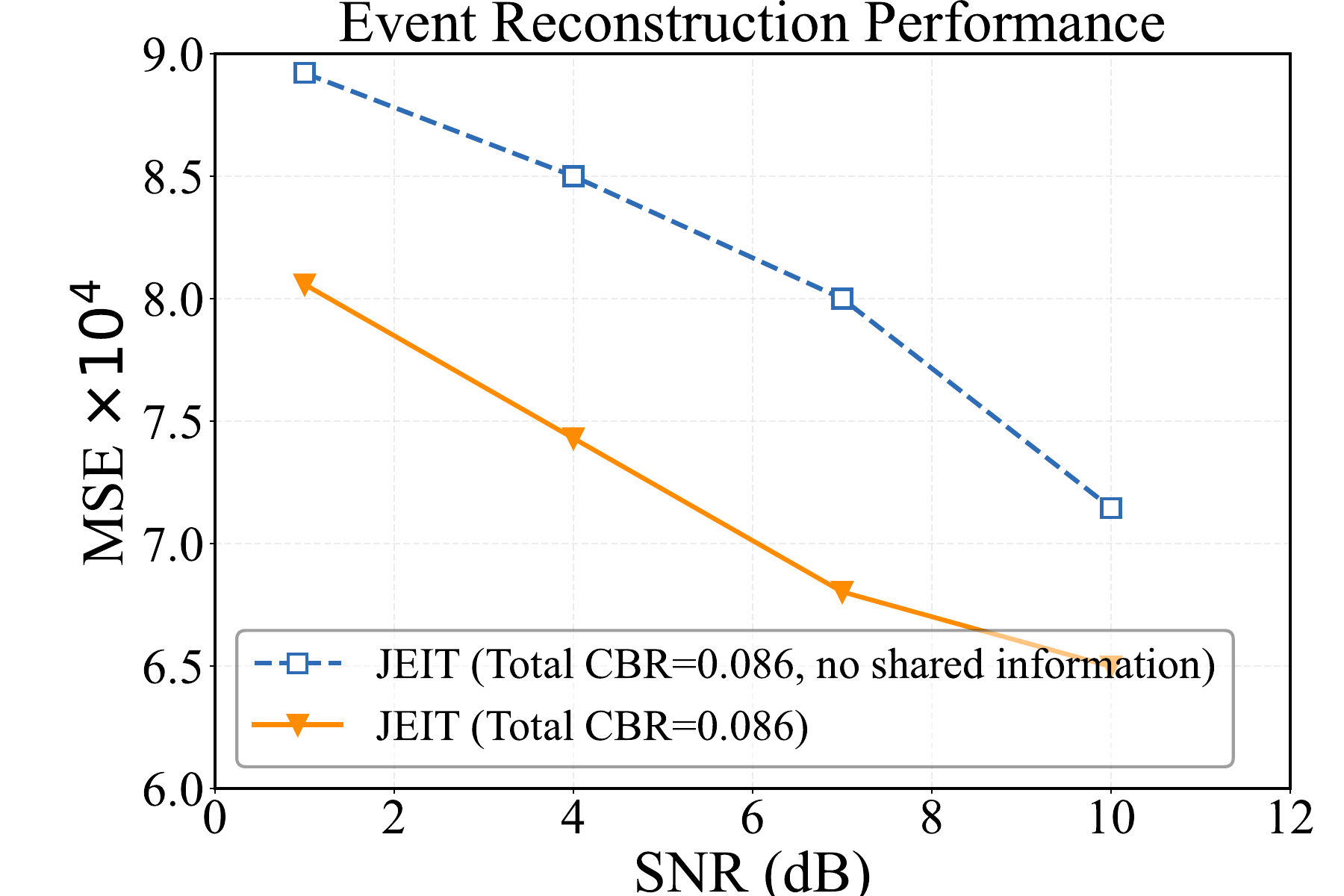}}
			\subfloat[]{\label{overhead1}\includegraphics[width=5.5cm]{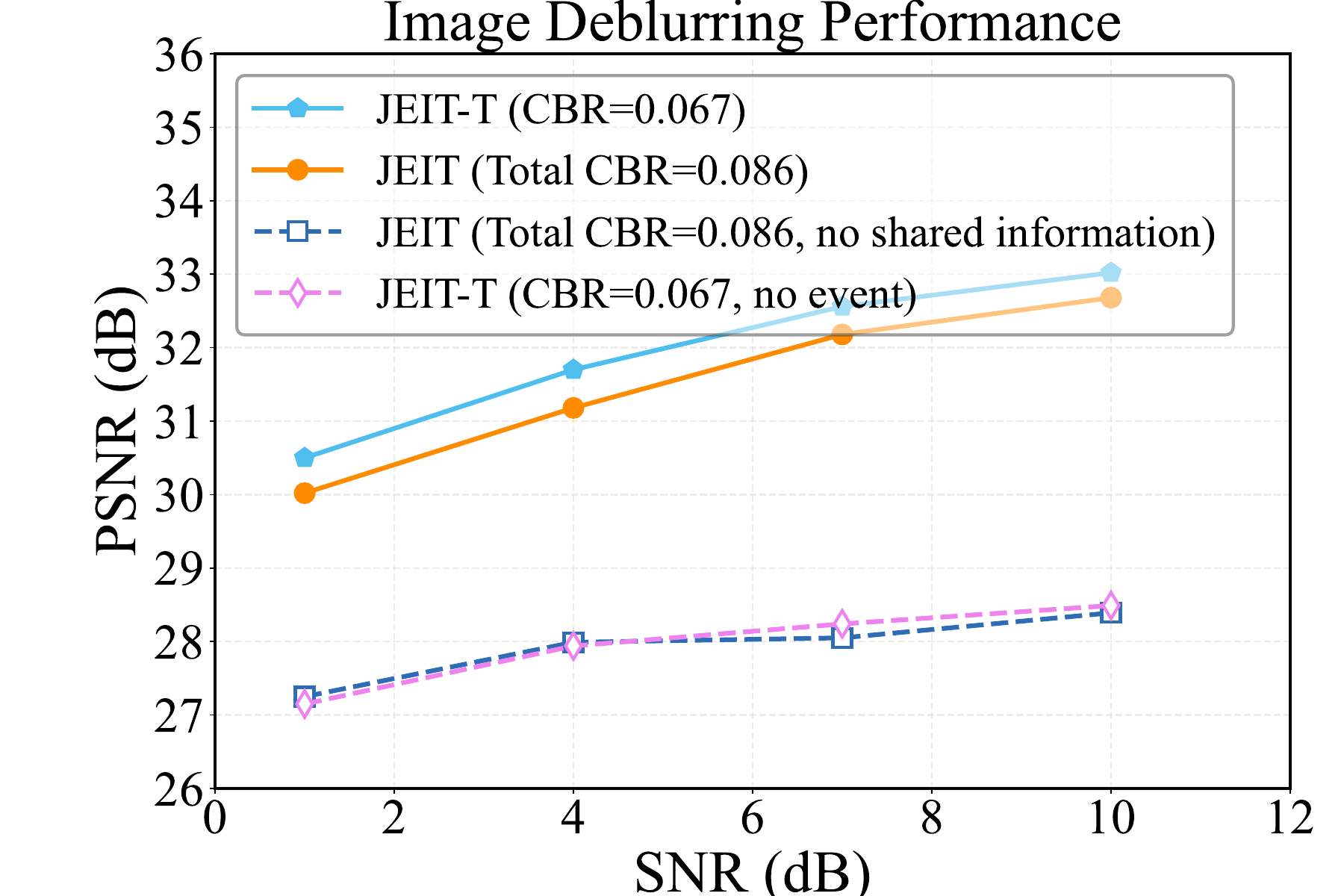}}
			\caption{The performance versus SNR over the AWGN channels of the total $\text{CBR}=0.16$ for image transmission, events transmission, and deblurring tasks.}
			\label{ablation}
		\end{centering}
	\end{figure*}
	In Figs. \ref{CBR1}(a) and \ref{CBR1}(b), we focus exclusively on the deblurring task, presenting results on the GoPro dataset across varying CBR levels under AWGN channels. The proposed method consistently achieves the best deblurring performance across all CBRs. 
	Notably, we can observe that when $\text{SNR} = 18\text{dB}$, the `EFNet + NTSCC' method shows minimal performance gain as CBR increases. This phenomenon likely arises from the limitations of either EFNet’s deblurring capability or NTSCC’s transmission capacity. In contrast, our method continues to show significant performance gains as CBR increases, with the performance gap further widening, highlighting the robustness of our approach under noise interference. Additionally, we evaluate performance on the REBlur dataset. Our scheme maintains strong deblurring performance, though the performance gap between methods is narrower compared to the GoPro dataset. This is primarily due to the lower resolution and milder blurring in REBlur images, which makes the transmission and deblurring task less challenging and results in relatively competitive performance across all evaluated methods.
	
	To further validate the effectiveness of the proposed methods, Fig. \ref{visualization2} showcases examples of deblurred images generated by different approaches (the `EFNet+DeepJSCC' method is omitted due to its significantly inferior performance compared to other benchmarks). The proposed methods consistently achieve superior visual quality, particularly at lower CBR levels. For instance, in the upper image, the reconstructions from `EFNet+SwinJSCC' and `EFNet+NTSCC' fail to preserve the sharp edge details of the white letter in the upper right corner, whereas JEIT-T accurately reconstructs these edges with high fidelity. Similarly, in the lower image, `EFNet+SwinJSCC' exhibits severe artifacts, while `EFNet+NTSCC' performs better due to its rate-allocation ability, which prioritizes areas with more intricate details. In contrast, the proposed methods excel at preserving texture and fine details, delivering sharp and faithful reconstructions. These results underscore the potential of the proposed methods for joint transmission and deblurring tasks.
	
	\subsection{Ablation Study}
	To assess the effectiveness of our proposed strategy for extracting shared and domain-specific information, we conduct a series of ablation studies. As illustrated in Fig. \ref{ablation}, transmitting only the independent features of the sources results in significant performance degradation across all tasks, with the most pronounced impact observed in deblurring. This is because deblurring relies more heavily on fused features, whereas image and event reconstruction can achieve satisfactory results using only domain-specific information. Additionally, to evaluate the importance of incorporating events for deblurring, we perform  an additional ablation study. As shown on the right side of Fig. \ref{ablation}(c), excluding events leads to a substantial decline in deblurring performance. This underscores both the limitations of RGB cameras in capturing high-speed scenarios and the critical role of events in supporting real-time tasks.

	\section{Conclusion} \label{SEC6}

	In this work, we proposed a joint event-image transmission and deblurring system. By leveraging the inherent correlation between events and images, we disentangled and transmitted both shared and domain-specific information. Specifically, we utilized two Bayesian networks to model the relationships among the inputs, the extracted information, and the outputs. We further analyzed the system via the lens of the information bottleneck principle and variational inference, deriving a tractable objective that balances reconstruction quality, deblurring performance, and transmission rate.
	The proposed framework was validated in a motion blur scenario, demonstrating its effectiveness. Overall, this paper presents a promising and efficient end-to-end approach for joint transmission and deblurring, achieving superior performance compared to existing methods.

	\begin{appendices}
		
		\section{Derivation of mutual information for base model} \label{app1}

		For maximization of $I(\mathbf{x}_1;\mathbf{\hat{y}}_1,\mathbf{\hat{y}}_2)$, the derivation process is similar to that of $I(\mathbf{x}_0;\mathbf{\hat{y}}_1,\mathbf{\hat{y}}_2)$ and is therefore omitted for brevity.
		
		For maximization of $I(\mathbf{t};\mathbf{\hat{y}}_0, \mathbf{\hat{y}}_1,\mathbf{\hat{y}}_2)$, we utilize a variational distribution $q(\mathbf{x}_1|\mathbf{\hat{y}}_1, \mathbf{\hat{y}}_2)$ to approximate the true posterior distribution:
		\begin{align}\label{distortion_t}
			&I(\mathbf{t};\mathbf{\hat{y}}_0, \mathbf{\hat{y}}_1,\mathbf{\hat{y}}_2) \notag \\
			&= \int p(\mathbf{\hat{y}}_0, \mathbf{\hat{y}}_1, \mathbf{\hat{y}}_2, \mathbf{t}) \log \frac{p(\mathbf{x}|\mathbf{\hat{y}}_0, \mathbf{\hat{y}}_1, \mathbf{\hat{y}}_2)}{p(\mathbf{t})} \,d\mathbf{t}\,d\mathbf{\hat{y}}_0\,d\mathbf{\hat{y}}_1\,d\mathbf{\hat{y}}_2 \notag \\
			&\geq \int p(\mathbf{\hat{y}}_0, \mathbf{\hat{y}}_1, \mathbf{\hat{y}}_2, \mathbf{t}) \log q(\mathbf{x}|\mathbf{\hat{y}}_0, \mathbf{\hat{y}}_1, \mathbf{\hat{y}}_2) \,d\mathbf{t}\,d\mathbf{\hat{y}}_0\,d\mathbf{\hat{y}}_1\,d\mathbf{\hat{y}}_2 \notag \\
			&+ H(\mathbf{t}),
		\end{align}
		where $H(\mathbf{t})$ is the entropy of deblurred image $\mathbf{t}$, only concerning the dataset, thus can be omitted during optimization. We assume $q(\mathbf{t}|\mathbf{\hat{y}}_0, \mathbf{\hat{y}}_1, \mathbf{\hat{y}}_2)$ to be a Gaussian distribution, which is given by:
		\begin{equation}
			q(\mathbf{t}|\mathbf{\hat{y}}_0, \mathbf{\hat{y}}_1, \mathbf{\hat{y}}_2) = \mathcal{N}(\mathbf{t}|\mathbf{\tilde{t}}, (2\lambda_t)^{-1}\mathbf{I}), \mathbf{\tilde{t}} = g_{\bm{\phi}_t}(\mathbf{\hat{y}}_0, \mathbf{\hat{y}}_1,\mathbf{\hat{y}}_2),
		\end{equation} 
		thus Eq. \eqref{distortion_t} can be simplified as:
		\begin{equation}
			I(\mathbf{t};\mathbf{\hat{y}}_1,\mathbf{\hat{y}}_2) \geq -\lambda_t \|\mathbf{t} - \mathbf{\tilde{t}}\|^2,
		\end{equation}
		which works out to be the MSE between $\mathbf{t}$ and $\mathbf{\tilde{t}}$. 
		
		For minimization of $I(\mathbf{x}_1;\mathbf{\hat{y}}_1)$, the derivation process is similar to that of $I(\mathbf{x}_0;\mathbf{\hat{y}}_0)$ and is therefore omitted for brevity.
		
		For minimization of $I(\mathbf{x}_0, \mathbf{x}_1; \mathbf{\hat{y}}_2)$, we utilize a parameterized distribution $q(\mathbf{\hat{y}}_2)$ to approximate the true marginal distribution:
		\begin{equation} 
			\begin{aligned}
				&I(\mathbf{x}_0,\mathbf{x}_1;\mathbf{\hat{y}}_2) \!=\!\! \int p(\mathbf{x}_0,\mathbf{x}_1;\mathbf{\hat{y}}_2) \log \frac{p(\mathbf{\hat{y}}_2|\mathbf{x}_0,\mathbf{x}_1)}{p(\mathbf{\hat{y}}_2)} \,d\mathbf{x}_0\,d\mathbf{x}_1\,d\mathbf{\hat{y}}_2 \\
				&\leq\! \int\! p(\mathbf{x}_0,\mathbf{x}_1;\mathbf{\hat{y}}_2) \big[\log p(\mathbf{\hat{y}}_2|\mathbf{x}_0,\mathbf{x}_1) \!-\! \log q(\mathbf{\hat{y}}_2)\big] d\mathbf{x}_0d\mathbf{x}_1d\mathbf{\hat{y}}_2,
			\end{aligned}
		\end{equation}
		where the first term can be expanded as:
		\begin{equation}
			\begin{aligned}
				p(\mathbf{\hat{y}}_2|\mathbf{x}_0,\mathbf{x}_1)   &=   \prod_{i}  \mathcal{U}(\hat{y}_{2_i}|y_{2_i} - \frac{1}{2},y_{2_i} + \frac{1}{2}), \\
				\mathbf{y}_2   &=   g_{e_2}(\mathcal{F}_{e_2}(\mathbf{x}_0, \mathbf{x}_1)),
			\end{aligned}
		\end{equation}
		which reflects the bit length required to encode $\mathbf{\hat{y}}_2$ into a bitstream.
		
		\section{Derivation of mutual information for enhanced model}
		For minimization of $I(\mathbf{\hat{z}}_1;\mathbf{\tilde{x}}_1)$, the derivation process is similar to that of $I(\mathbf{\hat{z}}_0;\mathbf{\tilde{x}}_0)$ and is therefore omitted for brevity.

		To minimize $I(\mathbf{\hat{z}}_2; \mathbf{x}_0, \mathbf{x}_1)$, we introduce a variational distribution $q(\mathbf{\hat{z}}_2)$,  leading to:
		\begin{equation} 
			\begin{aligned}
				&I(\mathbf{\hat{z}}_2; \mathbf{x}_0, \mathbf{x}_1) \!=\!\! \int p(\mathbf{x}_0,\mathbf{x}_1;\mathbf{\hat{z}}_2) \log \frac{p(\mathbf{\hat{z}}_2|\mathbf{x}_0,\mathbf{x}_1)}{p(\mathbf{\hat{z}}_2)} \,d\mathbf{x}_0\,d\mathbf{x}_1\,d\mathbf{\hat{z}}_2 \\
				&\leq\! \int\! p(\mathbf{x}_0,\mathbf{x}_1;\mathbf{\hat{z}}_2) \big[\log p(\mathbf{\hat{z}}_2|\mathbf{x}_0,\mathbf{x}_1) \!- \!\log q(\mathbf{\hat{z}}_2)\big] d\mathbf{x}_0d\mathbf{x}_1d\mathbf{\hat{z}}_2,
			\end{aligned}
		\end{equation}
		where the first term can be expanded as:
		\begin{equation}
			p(\mathbf{\hat{z}}_2|\mathbf{x}_0,\mathbf{x}_1) = \prod_{i} \mathcal{U}(\hat{z}_{2_i}|z_{2_i}-\frac{1}{2},z_{2_i}+\frac{1}{2}), \mathbf{z}_2 = h_{e_2}(\mathbf{y}_2),
		\end{equation}
		which reflects the bit length required to encode $\mathbf{\hat{z}}_2$.
		
		
		For $\big[-I(\mathbf{\hat{y}}_1; \mathbf{\hat{z}}_1) + \beta I(\mathbf{\hat{y}}_1; \mathbf{x}_1)\big]$, we set $\beta=1$ and optimize this term jointly following the similar procedure of optimizing $\big[-I(\mathbf{\hat{y}}_0; \mathbf{\hat{z}}_0) + \beta I(\mathbf{\hat{y}}_0; \mathbf{x}_0)\big]$.
		
		We jointly optimize $-I(\mathbf{\hat{y}}_2; \mathbf{\hat{z}}_2) + \beta I(\mathbf{\hat{y}}_2; \mathbf{x}_0, \mathbf{x}_1)$ with $\beta$ setting to $1$:
		\begin{align}
			&-I(\mathbf{\hat{y}}_2; \mathbf{\hat{z}}_2) + I(\mathbf{\hat{y}}_2; \mathbf{x}_0, \mathbf{x}_1) \notag \\
			&= \int p(\mathbf{\hat{y}}_2, \mathbf{\hat{z}}_2) \log \frac{p(\mathbf{\hat{y}}_2|\mathbf{\hat{z}}_2)}{p(\mathbf{\hat{y}}_2)} \,d\mathbf{\hat{y}}_2\,d\mathbf{\hat{z}}_2  \\
			&+ \int p(\mathbf{\hat{y}}_2, \mathbf{x}_0, \mathbf{x}_1) \log \frac{p(\mathbf{\hat{y}}_2|\mathbf{x}_0, \mathbf{x}_1)}{p(\mathbf{\hat{y}}_2)} \,d\mathbf{x}_0 \,d\mathbf{x}_1\,d\mathbf{\hat{y}}_2 \notag \\
			&\leq \int p(\mathbf{x}_0, \mathbf{x}_1, \mathbf{\hat{y}}_2, \mathbf{\hat{z}}_2) \big[-\log q(\mathbf{\hat{y}}_2|\mathbf{\hat{z}}_2) \notag \\
			&+ \log p(\mathbf{\hat{y}}_2|\mathbf{x}_0, \mathbf{x}_1)\big] \,d\mathbf{x}_0\,d\mathbf{x}_1\,d\mathbf{\hat{y}}_2\,d\mathbf{\hat{z}}_2. \notag
		\end{align}
		Here, $q(\mathbf{\hat{y}}_2|\mathbf{\hat{z}}_2)$ is a variational distribution introduced to approximate $p(\mathbf{\hat{y}}_2|\mathbf{\hat{z}}_2)$. The second term is a constant and can thus be omitted. For the first term, we model it as a Gaussian distribution convolved with a uniform distribution:
		\begin{equation} \label{rate_hyper_end}
			\begin{aligned}
				q(\mathbf{\hat{y}}_2|\mathbf{\hat{z}}_2) = \prod_{i} \left(\mathcal{N}(\hat{\mu}_{2_i},\hat{\sigma}_{2_i}^2) *\mathcal{U}(-\frac{1}{2}, \frac{1}{2})\right)(\hat{y}_{2_i}),
			\end{aligned}
		\end{equation}
		where $(\bm{\hat{\mu}}_2, \bm{\hat{\sigma}}_2) = h_{d_2}(\mathbf{\hat{z}}_2)$. This term reflects the bit length required to encode $\mathbf{\hat{y}}_2$ conditioned on $\mathbf{\hat{z}}_2$.

	\end{appendices}

	\bibliographystyle{IEEEtran}
	\bibliography{IEEEabrv,Reference}

\begin{thebibliography}{10}
\providecommand{\url}[1]{#1}
\csname url@samestyle\endcsname
\providecommand{\newblock}{\relax}
\providecommand{\bibinfo}[2]{#2}
\providecommand{\BIBentrySTDinterwordspacing}{\spaceskip=0pt\relax}
\providecommand{\BIBentryALTinterwordstretchfactor}{4}
\providecommand{\BIBentryALTinterwordspacing}{\spaceskip=\fontdimen2\font plus
\BIBentryALTinterwordstretchfactor\fontdimen3\font minus
  \fontdimen4\font\relax}
\providecommand{\BIBforeignlanguage}[2]{{%
\expandafter\ifx\csname l@#1\endcsname\relax
\typeout{** WARNING: IEEEtran.bst: No hyphenation pattern has been}%
\typeout{** loaded for the language `#1'. Using the pattern for}%
\typeout{** the default language instead.}%
\else
\language=\csname l@#1\endcsname
\fi
#2}}
\providecommand{\BIBdecl}{\relax}
\BIBdecl

\bibitem{EVJSCC}
P.~Yang, G.~Zhang, Y.~Cai, L.~Yu, and G.~Yu, ``Joint transmission and
  deblurring: A semantic communication approach using events,'' \emph{arXiv
  preprint arXiv:2501.09396}, 2025.

\bibitem{gallego2020event}
G.~Gallego, T.~Delbr{\"u}ck, G.~Orchard, C.~Bartolozzi, B.~Taba, A.~Censi,
  S.~Leutenegger, A.~J. Davison, J.~Conradt, K.~Daniilidis \emph{et~al.},
  ``Event-based vision: A survey,'' \emph{IEEE Trans. Pattern Anal. Mach.
  Intell}, vol.~44, no.~1, pp. 154--180, 2020.

\bibitem{glover2016event}
A.~Glover and C.~Bartolozzi, ``Event-driven ball detection and gaze fixation in
  clutter,'' in \emph{Proc. IEEE Int. Conf. Intell. Rob. Syst. (IROS)}, 2016,
  pp. 2203--2208.

\bibitem{yu2022learning}
L.~Yu, X.~Zhang, W.~Liao, W.~Yang, and G.-S. Xia, ``Learning to see through
  with events,'' \emph{IEEE Trans. Pattern Anal. Mach. Intell.}, vol.~45,
  no.~7, pp. 8660--8678, 2022.

\bibitem{kim2024frequency}
T.~Kim, H.~Cho, and K.-J. Yoon, ``Frequency-aware event-based video deblurring
  for real-world motion blur,'' in \emph{Proc. IEEE/CVF Conf. Comput. Vis.
  Pattern Recog. (CVPR)}, 2024, pp. 24\,966--24\,976.

\bibitem{xu2021motion}
F.~Xu, L.~Yu, B.~Wang, W.~Yang, G.-S. Xia, X.~Jia, Z.~Qiao, and J.~Liu,
  ``Motion deblurring with real events,'' in \emph{Proc. IEEE Int. Conf.
  Comput. Vis. (ICCV)}, 2021, pp. 2583--2592.

\bibitem{liang2024towards}
G.~Liang, K.~Chen, H.~Li, Y.~Lu, and L.~Wang, ``Towards robust event-guided
  low-light image enhancement: a large-scale real-world event-image dataset and
  novel approach,'' in \emph{Proc. IEEE/CVF Conf. Comput. Vis. Pattern Recog.
  (CVPR)}, 2024, pp. 23--33.

\bibitem{ldmic}
X.~Zhang, J.~Shao, and J.~Zhang, ``{LDMIC}: Learning-based distributed
  multi-view image coding,'' in \emph{Proc. Int. Conf. Learn. Represent.
  (ICLR)}, 2023.

\bibitem{AERprotocol}
K.~A. Boahen, ``Point-to-point connectivity between neuromorphic chips using
  address events,'' \emph{IEEE Trans. Circuits Syst. II, Analog Digit. Signal
  Process.}, vol.~47, no.~5, pp. 416--434, 2000.

\bibitem{maojin}
J.~Mao, K.~Xiong, M.~Liu, Z.~Qin, W.~Chen, P.~Fan, and K.~B. Letaief, ``A
  {GAN}-based semantic communication for text without {CSI},'' \emph{IEEE
  Trans. Wireless Commun.}, vol.~23, no.~10, pp. 14\,498--14\,514, 2024.

\bibitem{yufei2025}
Y.~Bo, S.~Shao, and M.~Tao, ``Deep learning-based superposition coded
  modulation for hierarchical semantic communications over broadcast
  channels,'' \emph{IEEE Trans. Commun.}, vol.~73, no.~2, pp. 1186--1200, 2025.

\bibitem{20254}
H.~Zhang, M.~Tao, Y.~Sun, and K.~B. Letaief, ``Improving learning-based
  semantic coding efficiency for image transmission via shared semantic-aware
  codebook,'' \emph{IEEE Trans. Commun.}, vol.~73, no.~2, pp. 1217--1232, 2025.

\bibitem{Mingyu_TCCN2022}
M.~Yang, C.~Bian, and H.-S. Kim, ``{OFDM}-guided deep joint source channel
  coding for wireless multipath fading channels,'' \emph{IEEE Trans. Cognit.
  Comm. Netw.}, vol.~8, no.~2, pp. 584--599, Jun. 2022.

\bibitem{guoxiansheng}
X.~Guo, G.~O. Boateng, H.~Si, Y.~Cao, Y.~Qiu, Z.~Lai, X.~Li, X.~Liu, and
  C.~Chen, ``Automated valet parking and charging: A novel collaborative
  {AI}-empowered architecture,'' \emph{IEEE Commun. Mag}, vol.~63, no.~1, pp.
  131--137, 2025.

\bibitem{Xiaojiao_IOTJ2024}
X.~Chen, J.~Wang, L.~Xu, J.~Huang, and Z.~Fei, ``A perceptually motivated
  approach for low-complexity speech semantic communication,'' \emph{IEEE
  Internet Things J.}, vol.~11, no.~12, pp. 22\,054--22\,065, 2024.

\bibitem{20253}
Y.~Liu, C.~Dong, H.~Liang, W.~Li, Z.~Bao, Z.~Zheng, X.~Xu, and P.~Zhang,
  ``Semantic-importance-aware reordering-enhanced semantic communication system
  with ofdm transmission,'' \emph{IEEE Internet Things J.}, vol.~12, no.~7, pp.
  7938--7954, 2025.

\bibitem{Guangyi_TCOM2024}
G.~Zhang, Q.~Hu, Z.~Qin, Y.~Cai, G.~Yu, and X.~Tao, ``A unified multi-task
  semantic communication system for multimodal data,'' \emph{IEEE Trans.
  Commun.}, vol.~72, no.~7, pp. 4101--4116, 2024.

\bibitem{cook2011interacting}
M.~Cook, L.~Gugelmann, F.~Jug, C.~Krautz, and A.~Steger, ``Interacting maps for
  fast visual interpretation,'' in \emph{Proc. Int. Jt. Conf. Neural Netw.
  (IJCNN)}, 2011, pp. 770--776.

\bibitem{bardow2016simultaneous}
P.~Bardow, A.~J. Davison, and S.~Leutenegger, ``Simultaneous optical flow and
  intensity estimation from an event camera,'' in \emph{Proc. IEEE/CVF Conf.
  Comput. Vis. Pattern Recog. (CVPR)}, 2016, pp. 884--892.

\bibitem{continuous}
C.~Scheerlinck, N.~Barnes, and R.~Mahony, ``Continuous-time intensity
  estimation using event cameras,'' in \emph{Proc. Asian Conf. Comput. Vis.
  (ACCV)}, 2018, pp. 308--324.

\bibitem{pan2019bringing}
L.~Pan, C.~Scheerlinck, X.~Yu, R.~Hartley, M.~Liu, and Y.~Dai, ``Bringing a
  blurry frame alive at high frame-rate with an event camera,'' in \emph{Proc.
  IEEE Int. Conf. Comput. Vis. (ICCV)}, 2019, pp. 6820--6829.

\bibitem{sun2022event}
L.~Sun, C.~Sakaridis, J.~Liang, Q.~Jiang, K.~Yang, P.~Sun, Y.~Ye, K.~Wang, and
  L.~V. Gool, ``Event-based fusion for motion deblurring with cross-modal
  attention,'' in \emph{Proc. Eur. Conf. Comput. Vis. (ECCV)}, 2022, pp.
  412--428.

\bibitem{wang2020event}
B.~Wang, J.~He, L.~Yu, G.-S. Xia, and W.~Yang, ``Event enhanced high-quality
  image recovery,'' in \emph{Proc. Eur. Conf. Comput. Vis. (ECCV)}, 2020, pp.
  155--171.

\bibitem{ni2015visual}
Z.~Ni, S.-H. Ieng, C.~Posch, S.~R{\'e}gnier, and R.~Benosman, ``Visual tracking
  using neuromorphic asynchronous event-based cameras,'' \emph{Neural Comput.},
  vol.~27, no.~4, pp. 925--953, 2015.

\bibitem{yu2023learning}
L.~Yu, B.~Wang, X.~Zhang, H.~Zhang, W.~Yang, J.~Liu, and G.-S. Xia, ``Learning
  to super-resolve blurry images with events,'' \emph{IEEE Trans. Pattern Anal.
  Mach. Intell.}, vol.~45, no.~8, pp. 10\,027--10\,043, 2023.

\bibitem{Huiqiang_TSP2021}
H.~Xie, Z.~Qin, G.~Y. Li, and B.-H. Juang, ``Deep learning enabled semantic
  communication systems,'' \emph{IEEE Trans. Signal Process.}, vol.~69, pp.
  2663--2675, Apr. 2021.

\bibitem{Eirina_TCCN2019}
E.~Bourtsoulatze, D.~Burth~Kurka, and D.~Gündüz, ``Deep joint source-channel
  coding for wireless image transmission,'' \emph{IEEE Trans. Cognit. Comm.
  Netw.}, vol.~5, no.~3, pp. 567--579, Sep. 2019.

\bibitem{zhang2025HJSCC}
G.~Zhang, H.~Li, Y.~Cai, Q.~Hu, G.~Yu, and Z.~Qin, ``Progressive learned image
  transmission for semantic communication using hierarchical {VAE},''
  \emph{IEEE Trans. Cognit. Comm. Netw.}, pp. 1--1, to appear, 2025,
  10.1109/TCCN.2025.3546935.

\bibitem{lamosc}
Y.~Zhao, Y.~Yue, S.~Hou, B.~Cheng, and Y.~Huang, ``La{M}o{SC}: Large language
  model-driven semantic communication system for visual transmission,''
  \emph{IEEE Trans. Cognit. Comm. Netw.}, vol.~10, no.~6, pp. 2005--2018, 2024.

\bibitem{time2020}
N.~Khan, K.~Iqbal, and M.~G. Martini, ``Time-aggregation-based lossless video
  encoding for neuromorphic vision sensor data,'' \emph{IEEE Internet Things
  J.}, vol.~8, no.~1, pp. 596--609, 2020.

\bibitem{Jincheng_JSAC2022}
J.~Dai, S.~Wang, K.~Tan, Z.~Si, X.~Qin, K.~Niu, and P.~Zhang, ``Nonlinear
  transform source-channel coding for semantic communications,'' \emph{IEEE J.
  Select. Areas Commun.}, vol.~40, no.~8, pp. 2300--2316, 2022.

\bibitem{Bayesian}
Z.~Dang, M.~Luo, C.~Jia, G.~Dai, J.~Wang, X.~Chang, and J.~Wang, ``Disentangled
  representation learning with transmitted information bottleneck,'' \emph{IEEE
  Trans. Circuits Syst. Video Technol.}, 2024.

\bibitem{multiinformation}
M.~Studen{\`y} and J.~Vejnarov{\'a}, ``The multiinformation function as a tool
  for measuring stochastic dependence,'' \emph{Learning in graphical models},
  pp. 261--297, 1998.

\bibitem{multivariate2013}
N.~Slonim, N.~Friedman, and N.~Tishby, ``Multivariate information bottleneck,''
  \emph{Neural Comput.}, vol.~18, no.~8, pp. 1739--1789, 2006.

\bibitem{balle_arxiv2018}
J.~Ball{\'e}, D.~Minnen, S.~Singh, S.~J. Hwang, and N.~Johnston, ``Variational
  image compression with a scale hyperprior,'' in \emph{Proc. Int. Conf. Learn.
  Represent. (ICLR)}, 2018.

\bibitem{vmamba}
Y.~Liu, Y.~Tian, Y.~Zhao, H.~Yu, L.~Xie, Y.~Wang, Q.~Ye, J.~Jiao, and Y.~Liu,
  ``{VM}amba: Visual state space model,'' in \emph{Proc. Adv. Neural Inf.
  Process. Syst. (NeurIPS)}, 2025, pp. 103\,031--103\,063.

\bibitem{gopro}
S.~Nah, T.~Hyun~Kim, and K.~Mu~Lee, ``Deep multi-scale convolutional neural
  network for dynamic scene deblurring,'' in \emph{Proc. IEEE/CVF Conf. Comput.
  Vis. Pattern Recog. (CVPR)}, 2017, pp. 3883--3891.

\bibitem{esim}
H.~Rebecq, D.~Gehrig, and D.~Scaramuzza, ``{ESIM}: an open event camera
  simulator,'' in \emph{Proc. Conf. Robot Learn. (CoRL)}, 2018, pp. 969--982.

\bibitem{ADJSCC}
J.~Xu, B.~Ai, W.~Chen, A.~Yang, P.~Sun, and M.~Rodrigues, ``Wireless image
  transmission using deep source channel coding with attention modules,''
  \emph{IEEE Trans. Circuits Syst. Video Technol.}, vol.~32, no.~4, pp.
  2315--2328, Apr. 2022.

\bibitem{SwinJSCC}
K.~Yang, S.~Wang, J.~Dai, X.~Qin, K.~Niu, and P.~Zhang, ``Swin{JSCC}: Taming
  swin transformer for deep joint source-channel coding,'' \emph{IEEE Trans.
  Cognit. Comm. Netw.}, vol.~11, no.~1, pp. 90--104, 2025.

\end{thebibliography}
	
\end{document}